\documentclass[a4paper]{cas-sc}

\usepackage[utf8]{inputenc}
\usepackage[authoryear]{natbib}

\usepackage{amsmath,amssymb,amsfonts}
\usepackage{graphicx}
\usepackage{subcaption}
\usepackage{caption}
\usepackage{booktabs}
\usepackage{multirow}
\usepackage{array}
\usepackage{tabularx}
\usepackage{svg}
\usepackage{float}
\usepackage{placeins}
\usepackage{hyperref}
\usepackage[dvipsnames]{xcolor}
\usepackage{longtable}
\usepackage{rotating}
\usepackage{makecell}

\usepackage{siunitx}
\begin{document}

\let\WriteBookmarks\relax
\def\floatpagepagefraction{1}
\def\textpagefraction{.001}

\shorttitle{}
\shortauthors{}


\title[mode=title]{Physics-Constrained Autoencoder for Interpretable Health Representation Learning in Turbofan Engines}


\author[1]{Leonardo Baldo}
\fnmark[†]
\cormark[1]
\ead{*Corresponding Author leonardo.baldo@zhaw.ch}

\author[2]{Jokin Alcibar~Ibarzabal}
\fnmark[†]
\ead{j.alcibar@mondragon.edu}

\author[3]{Daniel Cisneros~Acevedo}
\ead{D.I.CisnerosAcevedo@tudelft.nl}

\author[4,5]{Kai Goebel}
\ead{kai.goebel@fragumglobal.com}

\author[1,3]{Manuel Arias~Chao}
\ead{aria@zhaw.ch}

\address[1]{Institute for Data Science, Zurich University of Applied Science, Winterthur, 8401, Switzerland}

\address[2]{Computer Science Department, Electronics, Mondragon University, Loramendi, 4, Mondragón, 20500, Spain}

\address[4]{Fragum Global, Mountain View, CA 94040, USA}

\address[5]{Luleå University of Technology, Luleå, 971 87, Sweden}

\address[3]{Faculty of Aerospace Engineering, Delft University of Technology, HS 2926 Delft, The Netherlands}

\nonumnote{†\, These authors contributed equally to this work and share first authorship.}


\maketitle

\begin{abstract}
Health Indices (HIs) summarize degradation from operational data and support diagnostic and prognostic tasks in Prognostics and Health Management (PHM). Their usefulness depends on capturing degradation and interpreting system condition. Autoencoders learn such representations from unlabelled monitoring data; however, when training is driven only by reconstruction, the physical meaning of latent variables remains undefined, limiting their interpretation and diagnostic value.
This work proposes a physics-constrained autoencoder that embeds system knowledge as an inductive bias in the decoder architecture and as a learning bias in the loss function. The decoder contains differentiable physical equations that receive latent variables and operating conditions as inputs, structurally guiding each latent variable to represent a predefined physical quantity and enabling component-level analysis before aggregation into a system-level HI.
The methodology is benchmarked on the N-CMAPSS dataset against alternative representations. Reconstruction reaches an $R^2$ of 0.9993 and an MAE of 0.0036, comparable to fully data-driven models. After operating-condition correction, the inferred physical quantities track degradation, and their aggregation produces an HI with a Spearman correlation of 0.9791 with the ground truth, an RMSE of 0.0286, and an MAE of 0.0233. The methodology is further evaluated using a single model trained across four datasets representing distinct degradation scenarios. This model preserves component-level signatures that distinguish compression-side from expansion-side degradation without diagnostic labels.
Finally, the proposed framework enables degradation trends to be visualized on component maps, providing a health representation in a form familiar to engine operators and domain experts.

\end{abstract}


\section{Introduction}
\label{sec:intro}

Prognostics and Health Management (PHM) has progressively reshaped the way operational data are processed and used in industrial settings \cite{ZIO2022108119}. The increasing availability of sensors, embedded computing resources, and advanced data-processing methodologies has made it possible to move beyond traditional maintenance strategies, enabling informed asset management based on predictive decision-making \cite{8842589}. In this context, PHM provides a structured framework for transforming raw monitoring data into actionable information supporting fault detection, diagnosis, prognosis, and health-aware asset management \cite{HU2022108063}. The main objectives of PHM can therefore be formulated as (i) assessing the current condition of an asset through diagnostics, (ii) anticipating its future degradation through prognostics, and (iii) supporting maintenance and operational decisions through health management \cite{Jardine2006CBM}.

The emergence of PHM is commonly associated with the aerospace and defence sectors, where the need for safer, more available, and more cost-effective systems has always been particularly demanding \cite{Hess2005BigP, jimenez2026conceptual}. Aircraft systems and mission-critical platforms, including turbofan engines \cite{Xu2026TurbofanHealthAssessment}, missile guidance systems \cite{Hu2024MissileFaultDetection}, and flight-control actuators \cite{Baldo2025FlightControlPHMReview}, have motivated the development of integrated monitoring capabilities able to detect incipient faults, estimate Remaining Useful Life (RUL), and support logistics and maintenance planning. Since then, PHM has expanded well beyond its initial aeronautic domain, spanning a plethora of applications. Today, PHM methodologies are investigated and applied in several industrial sectors, including actuation systems \cite{bertolino2023design}, space
\cite{aimasso2022analysis}, automotive applications \cite{BUSTOS2026114449}, batteries \cite{alcibar2026bayesian}, and process industries \cite{HU2022108063}.

Over more than two decades of development, PHM has continuously absorbed methodological advances from statistics, signal processing, control theory, Machine Learning (ML), and Deep Learning (DL). Indeed, since their inception, classical ML algorithms have enabled patterns to be extracted from monitoring data with less dependence on manually designed diagnostic rules \cite{Lei2020MLFaultDiagnosis}. More recently, DL has further extended these capabilities by allowing models to learn representations directly from high-dimensional monitoring data and extract health tracking variables \cite{Fink2020DeepPHM}.

A central concept in this context is the Health Index (HI), which is commonly introduced as a representation of the degradation state of an asset measured through a normalized version of a health indicator. In other words, it can be defined as a compact representation of asset condition designed to track degradation progression and support downstream diagnostic and prognostic tasks. 
In practical applications, HIs have been obtained through different strategies, including the selection of degradation-sensitive features, the use of model residuals, the fusion of multiple condition-monitoring signals, and the learning of latent degradation representations from data \cite{Zhou2022HIReview, chen2025constructing,ge2025logcumulative}.

Among data-driven approaches, autoencoders have become a widely adopted solution for unsupervised health representation learning. Their encoder--decoder structure allows high-dimensional monitoring data to be compressed into a compact latent representation and reconstructed from it \cite{Bengio2013RepresentationLearning}. In PHM, the latent variables or a post-processed function of the latent space (e.g., a latent dynamical model) can then be used as an indicator of degradation and, consequently, as an HI. This makes autoencoder-based approaches particularly attractive when health-state labels are unavailable, run-to-failure data are limited, or the degradation state is only indirectly observed through sensor measurements, a combination frequently encountered in industrial applications \cite{GonzalezMuniz2022LatentHI,DePater2023LSTMAE}.

Despite the intuitive appeal of using a latent representation as a degradation descriptor, a standard autoencoder is designed and optimized primarily for reconstruction. As a consequence, its latent coordinates are not guaranteed to correspond to physically meaningful degradation mechanisms. In fact, they may encode operating conditions, sensor drift, noise, mission-related trends, or dataset-specific correlations that help reconstruct the input data without necessarily describing the health state of the asset.
This limitation is especially significant from a user perspective, considering the vital role of HI as key metrics that operators rely on to evaluate asset condition and determine the necessity of intervention. When an HI is obtained from a black-box latent space, its physical meaning may remain unclear even when the reconstruction performance of the model is satisfactory. Moreover, even when a learned latent representation tracks degradation accurately, it may provide little information about the physical source of that degradation or the component responsible for the observed change. Such ambiguities can reduce trustworthiness, especially in safety-critical applications where operators need to understand why a model indicates degradation and whether this indication is consistent with known system behaviour. Obtaining HIs that are predictive, interpretable, physically consistent, and usable by domain experts is therefore critical for the further adoption of PHM systems in industrial settings \cite{Nor2021XAI}.

Against this backdrop, Physics-Informed Machine Learning (PIML) offers a possible way to address this issue by constraining the learning process with prior knowledge about the system \cite{wu2024physicsinformed}. In PHM, this prior knowledge can take the form of governing equations, simplified physical models, degradation laws, monotonicity constraints, expert knowledge, or structural assumptions about the relationships among operating conditions, health state, and measured variables \cite{Deng2023PIML,Li2024PhysicsInformedRULReview}. More specifically, the mechanisms through which prior knowledge is introduced into an ML model are commonly organized into three categories \cite{Bajarunas2023PIHI}: observational bias, inductive bias, and learning bias \cite{Karniadakis2021PIML, Fink2026PhysicsPartI}. 
Recent studies have shown that physical information can guide HI construction, regularize degradation trajectories, and improve the consistency of learned representations under limited data and varying operating conditions \cite{Bajarunas2023PIHI,Bajarunas2024HIKnowledge}. Recent physics-derived decoders and constraint-guided optimization strategies have further demonstrated that physically meaningful quantities and degradation requirements can be embedded into autoencoder-based HI learning, reinforcing the relevance of these architectures and inherently motivating the present work \cite{Huang2024PhysicsInformedAE,Tefera2025ConstraintGuidedHI, plonczak2026physics}.

However, existing physics-guided HI approaches predominantly constrain the evolution of a system-level health representation or embed physical information into low-dimensional latent variables. They do not generally provide a multidimensional, component-resolved latent representation in which individual coordinates are structurally associated with physical quantities of different components while jointly reconstructing the monitored system response. Consequently, a gap remains between learning a degradation-sensitive HI and obtaining a health representation that can also provide physically structured information about where and how degradation manifests within the system in a relatively high dimensionality setting.

This work addresses this gap by proposing a physics-decoder autoencoder for unsupervised and interpretable health representation learning in PHM. Within the proposed formulation, the encoder maps measured variables and operating conditions into a latent representation, while the decoder reconstructs the measurements through a structured model inspired by the physical organization of the monitored system. The physical prior is then of an equation-based type and physical knowledge is introduced primarily as an inductive bias and secondly with a learning bias. The differentiable decoder is structured according to physics and component equations, restricting the learned latent representation to influence the reconstructed measurements through predefined physical modifiers. As a result, embedding the physical equations in the decoder forces each latent variable to represent a predefined physical quantity or degradation effect, thereby giving the latent space a direct physical interpretation. The latent variables reach the sensor outputs through these physically guided equations and can therefore be interpreted as degradation modifiers affecting component and system performance. This provides a component-aware physical interpretation of the latent representation, subject to the observability and identifiability of the corresponding quantities.
The framework is finally benchmarked and validated on several N-CMAPSS turbofan datasets \cite{AriasChao2021NCMAPSS}. 
The main contributions of this work are as follows:
\begin{itemize}
\item a general physics-decoder autoencoder framework for unsupervised and interpretable health representation learning in PHM, which embeds differentiable system equations in the decoder and physics-based residuals in the loss function as complementary inductive and learning biases;

\item a component-aware formulation that forces each latent variable in a 15-dimensional space to represent a predefined physical quantity and act as a degradation-sensitive modifier within the corresponding equations, enabling component-level examination before system-level aggregation;

\item a strategy that removes the operating-condition contribution from the learned physical latent variables and combines their most informative degradation trends into a system-level health index;

\item an extensive N-CMAPSS evaluation spanning four degradation scenarios, including benchmarks against alternative representations, an independent consistency assessment using two high-fidelity turbofan simulators, an analysis of latent-space observability, visualization through component maps, and separation of compressor-side from turbine-side degradation without training a diagnostic classifier.
\end{itemize}



\section{Health Indices and Positioning of the Proposed Approach}
\label{sec:health_indicators}

Health Indices (HIs) provide the interface between condition-monitoring data and health-informed decisions. For prognostic applications, an HI should follow degradation consistently and retain information relevant to future steps in the PHM pipeline, such as health assessment, RUL estimation, and maintenance decision-making. Monotonicity, trendability, robustness, prognosability, and consistency across units are therefore commonly used to evaluate HI quality \cite{Coble2009OptimalPrognosticParameters,Lei2018MachineryHealthPrognostics}.

The evolution of HI construction reflects the broader methodological development of PHM. In the next paragraph a non-exhaustive list of the major contributions are examined and the results are reported in Table~\ref{tab:hi_generation_positioning}, which summarizes the main HI construction families and clarifies the positioning of the proposed formulation. A more detailed overview of the most recent methodologies to extract HI is provided by Bajarunas et al. \cite{Bajarunas2024HIKnowledge}. 

Early approaches relied on degradation-sensitive statistics and signal-processing features selected through knowledge of the monitored component and its failure mechanisms (e.g. kurtosis, spectral amplitudes, and energy-related quantities) \cite{Lei2018MachineryHealthPrognostics,Zhou2022HIReview}. Model-based diagnostic approaches subsequently introduced a more explicit description of system behaviour, allowing asset condition to be represented through residuals or estimated parameters \cite{Jardine2006CBM,Coble2009OptimalPrognosticParameters}.
DL methods shifted HI construction towards representations learned directly from monitoring data. Autoencoders are particularly suitable for this purpose because they compress unlabeled and high-dimensional measurements while preserving the information required for their reconstruction. This is why they are among the most popular architectures for extracting HI from unlabeled operational data in an unsupervised way. For instance, González-Muñiz et al.\ construct an HI from the discrepancy between the latent representations of the measured and reconstructed signals \cite{GonzalezMuniz2022LatentHI}. De Pater and Mitici aggregate the reconstruction errors of an operating-condition-informed LSTM autoencoder \cite{DePater2023LSTMAE}. These approaches provide data-driven measures of drift from learned system behaviour. However, their physical interpretation remains limited because the reconstruction objective does not assign specific meanings to the latent coordinates. Consequently, the latent space can combine degradation with operating conditions, sensor characteristics, and other sources of variability that contribute to reconstruction.
This limitation has motivated the introduction of prior knowledge into health representation learning. One of the first examples is the one by Bajarunas et al., who encode structural assumptions concerning the relationships among operating conditions, degradation, and measurements, and introduce constraints that promote degradation-consistent latent trajectories \cite{Bajarunas2023PIHI,Bajarunas2024HIKnowledge}. Tefera et al.\ guide the estimation of a scalar HI through monotonicity, boundary, and signal-energy consistency constraints \cite{Tefera2025ConstraintGuidedHI}. These contributions shape the behaviour of the learned representation through degradation-related knowledge, while retaining data-driven mappings between the latent variables and the reconstructed measurements.
A further development consists of embedding physical knowledge within the autoencoder decoder itself. The closest methodological reference is the physics-informed autoencoder proposed by Huang et al.\ \cite{Huang2024PhysicsInformedAE}. Their encoder estimates a scalar latent HI with a predefined energy-oriented interpretation, corresponding to battery state of health or electromechanical-actuator efficiency. Physics-derived feature terms are introduced in the decoder and combined through trainable linear or neural layers to reconstruct the monitored output. The reconstruction process therefore links the scalar latent variable to a defined measure of overall system performance. A related formulation is proposed by Plonczak et al.\ \cite{plonczak2026physics}, who couple a convolutional encoder with an equivalent-circuit model whose reconstruction equations force the latent variables to represent effective capacity and internal resistance for battery packs.

Building on this line of research, the present work introduces an equation-based autoencoder for multidimensional and physically structured health representation learning. The methodological formulation assigns the latent coordinates to modifiers of named physical quantities and embeds these modifiers within a coupled system of differentiable equations. A system-level HI can subsequently be extracted from this structured representation while preserving information about its underlying physical contributions. Unlike Huang et al., the equations directly constitute the decoder and transform the latent representation and operating conditions into reconstructed measurements, with no neural mapping after the equations. This more constrained structure allows the latent space to retain several degradation contributions and associates each coordinate with the physical quantity it modifies. Moreover, Plonczak et al. constrain two latent variables through a battery equivalent-circuit model. We extend this idea to multicomponent systems through a 15-dimensional, component-resolved latent space and a decoder assembled from component equations instead of a model. The framework also removes operating-condition effects before HI construction and evaluates latent observability and component-level diagnosis across four degradation scenarios in different areas of the monitored asset.

\begin{table*}[t]
\centering
\caption{Representative HI construction strategies and positioning of the proposed component-resolved physics-decoder autoencoder.}
\label{tab:hi_generation_positioning}
\renewcommand{\arraystretch}{1.08}
\setlength{\tabcolsep}{3pt}
\scriptsize
\begin{tabularx}{\textwidth}{p{2.45cm} X X p{2.45cm}}
\toprule
\textbf{Method family} &
\textbf{HI construction and knowledge integration} &
\textbf{Interpretability and scope} &
\textbf{Representative refs.} \\
\midrule

Expert-designed HIs &
Statistical or signal-processing quantities selected from known degradation signatures. &
Explicit meaning, but dependent on operating conditions, failure mode, and feature selection. &
\cite{Lei2018MachineryHealthPrognostics,Zhou2022HIReview} \\

Model- and residual-based HIs &
Residuals or estimated parameters based in regression based models. Knowledge enters through the model structure and reference behaviour. &
Interpretation follows from the model and residual definition; performance depends on reference quality and model fidelity. &
\cite{Jardine2006CBM,Coble2009OptimalPrognosticParameters} \\

Autoencoder residual HIs &
Input or latent reconstruction errors or sensor reconstruction losses, possibly conditioned. &
Quantify departure from learned behaviour without identifying the physical source of the change. &
\cite{GonzalezMuniz2022LatentHI,DePater2023LSTMAE} \\

Knowledge-constrained learned HIs &
Latent representations or HI outputs constrained by causal structure, monotonicity, temporal ordering, or boundary conditions. &
Degradation-consistent trajectories; meaning depends on the imposed constraints. &
\cite{Bajarunas2023PIHI,Bajarunas2024HIKnowledge,Tefera2025ConstraintGuidedHI} \\

Hybrid physics-decoder HIs &
A scalar latent space with predefined meaning enters physics-derived decoder terms, possibly combined by later trainable layers. &
Provides a system-level physical interpretation, but estimates a single HI and requires a system-specific model. &
\cite{Huang2024PhysicsInformedAE, plonczak2026physics} \\

\textbf{Proposed component-resolved physics-decoder AE} &
\textbf{A latent vector modifies specific quantities within coupled equations that directly reconstruct multiple measurements.} &
\textbf{Coordinates are structurally associated with component and performance degradation effects; interpretability depends on model fidelity and latent identifiability/observability.} &
\textbf{This work} \\

\bottomrule
\end{tabularx}
\end{table*}

\section{Methodology: Physics-Constrained Autoencoder Formulation}
\label{sec:physics_ae_formulation}

The proposed methodology aims to infer a structured latent representation whose coordinates are associated with predefined physical quantities and whose evolution can subsequently be examined for health assessment. Its application requires prior knowledge of the monitored system in the form of differentiable physical relations linking the operating conditions, the relevant component parameters, and the measured system response. This knowledge defines the inductive and learning bias introduced into the autoencoder and indirectly determines the physical meaning assigned to the latent coordinates.

Consider a monitored system described at time $t$ by a vector of measured variables
$\mathbf{x}_t\in\mathbb{R}^{n_y}$ and a vector of operating conditions
$\mathbf{w}_t\in\mathbb{R}^{n_w}$. The objective is to infer a latent representation
$\mathbf{z}_t\in\mathbb{R}^{n_z}$ from which the measured system response can be reconstructed. The resulting latent representation is expected to retain information on the physical state of the monitored components. Since physical component quantities may also vary with the operating regime, a subsequent operating-condition correction may be required before their temporal evolution can be interpreted in terms of degradation.

\subsection{Conventional Conditional Autoencoder Formulation}
A conventional conditional autoencoder consists of an encoder
$\mathcal{E}_{\phi}:\mathbb{R}^{n_y+n_w}\rightarrow\mathbb{R}^{n_z}$ and a neural decoder
$\mathcal{D}_{\theta}:\mathbb{R}^{n_z+n_w}\rightarrow\mathbb{R}^{n_y}$, both composed of trainable neural layers:
\begin{align}
\mathbf{z}_t
&=
\mathcal{E}_{\phi}\!\left(\mathbf{x}_t,\mathbf{w}_t\right),
\label{eq:conditional_encoder}
\\
\widehat{\mathbf{x}}_t
&=
\mathcal{D}_{\theta}\!\left(\mathbf{z}_t,\mathbf{w}_t\right),
\label{eq:conditional_neural_decoder}
\end{align}

where $\phi$ and $\theta$ denote the trainable parameters of the encoder and decoder, respectively. Without loss of generality, the domains of the two mappings are written assuming that the measured variables and operating conditions are concatenated at the encoder input, while the latent representation and operating conditions are concatenated at the decoder input. Alternative conditioning mechanisms, including feature-wise modulation \cite{Perez2018} or the injection of $\mathbf{w}_t$ only at selected layers, can be adopted without changing the general formulation.

Providing $\mathbf{w}_t$ to both mappings allows the encoder to infer a latent representation conditional on the current operating regime and allows the decoder to reconstruct the measurements at the same operating point. However, when training is based exclusively on reconstruction, no constraint requires the latent representation to be independent of $\mathbf{w}_t$. The operating-condition information may therefore remain encoded in $\mathbf{z}_t$, even when it is explicitly supplied to the decoder. Moreover, no predefined physical meaning is assigned to the individual latent coordinates, since the neural architecture does not specify their correspondence with the physical quantities governing the system response.

\subsection{Physics-Constrained Decoder}

\begin{figure*}
\centering
\includegraphics[width=0.95\textwidth]{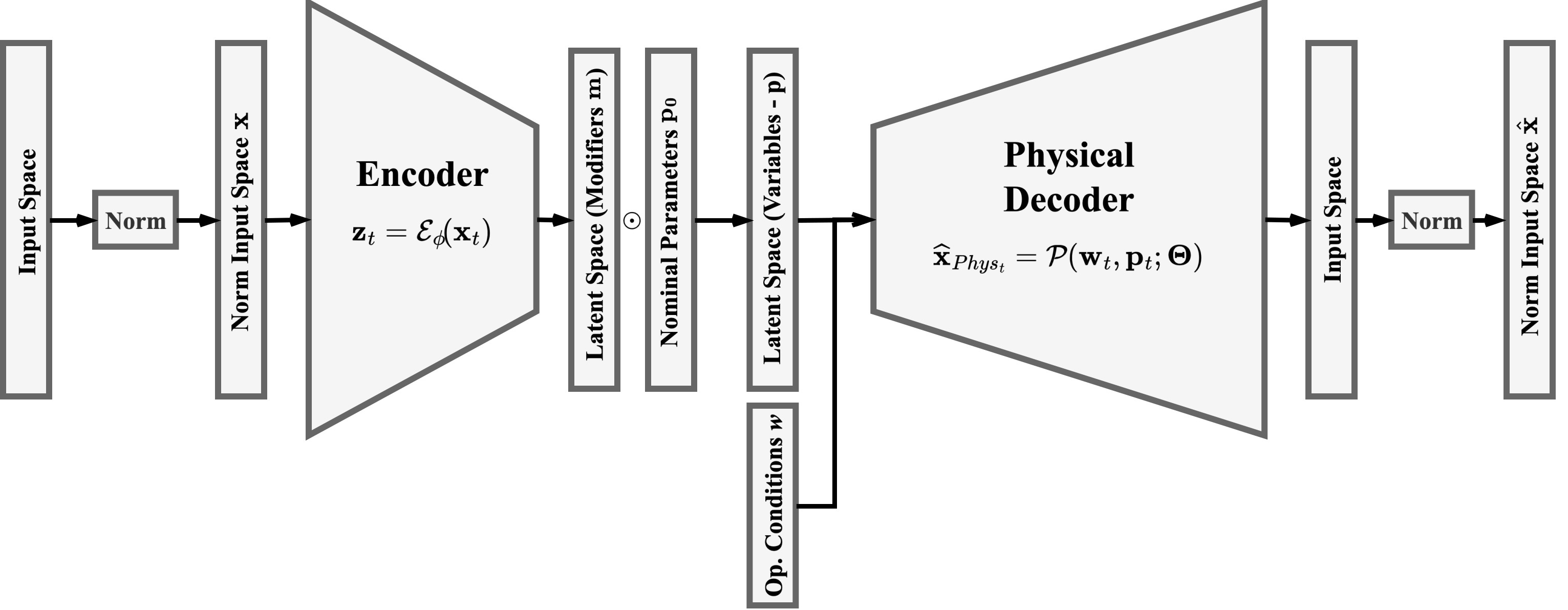}
\caption{Generic architecture of the proposed physics-constrained autoencoder. The encoder maps the normalized measurements $\mathbf{x}_t$ into a structured latent representation $\mathbf{z}_t$. Each latent variable is associated with a predefined physical quantity and is propagated, together with the operating conditions $\mathbf{w}_t$, through the differentiable physical decoder to reconstruct the measured variables $\hat{\mathbf{x}}_t$.}
\label{fig:generic_model}
\end{figure*}

In the proposed formulation, the neural decoder $\mathcal{D}_{\theta}$ is replaced by a differentiable physical model $\mathcal{P}(\cdot)$, as shown in Figure \ref{fig:generic_model}. Each latent coordinate is associated with a physical quantity selected through an analysis of the system structure, its governing equations, and the degradation mechanisms of interest. This analysis can be carried out for instance as a sensitivity analysis of the model to identify which variables should be placed in the latent space. The encoder output can be first transformed into a vector of dimensionless physical modifiers:
$\mathbf{m}_t
=
\mathcal{R}\!\left(\mathbf{z}_t\right)
=
\begin{bmatrix}
r_1\!\left(z_{1,t}\right) &
\cdots &
r_{n_z}\!\left(z_{n_z,t}\right)
\end{bmatrix}^{\top},$ where $\mathcal{R}(\cdot)$ is a fixed component-wise transformation and
$r_j(\cdot)$ maps the $j$-th latent coordinate to the admissible range of its assigned physical modifier. The form of each transformation is defined before training according to the domain and expected variation of the corresponding physical quantity. For example, a strictly positive modifier can be defined as
$m_{k,t}
=
r_k\!\left(z_{k,t}\right)
=
e^{0.05\,z_{k,t}}$ for a generic component $k$.
The transformations introduce no additional trainable parameters and ensure positivity.

Let $\mathbf{p}_{0}\in\mathbb{R}^{n_z}$ denote the nominal (reference) values of the physical quantities associated with the latent coordinates. Their time-dependent values are defined as $\mathbf{p}_t
=
\mathbf{p}_{0}\odot\mathbf{m}_t,$ with $\odot$ that denotes the Hadamard product. The vector $\mathbf{p}_t$ collects the physical parameters through which changes in the monitored components affect the system response. 
The reconstructed measurements are obtained by evaluating the physical model under the observed operating conditions:
\begin{equation}
\widehat{\mathbf{x}}_t
=
\mathcal{P}\!\left(
    \mathbf{w}_t,
    \mathbf{p}_t;
    \boldsymbol{\Theta}
\right),
\label{eq:physical_decoder}
\end{equation}
where $\boldsymbol{\Theta}$ contains the fixed (i.e., non-trainable) physical constants and nominal model quantities that are not inferred from the latent representation. The function
$\mathcal{P}(\cdot)$ comprises the governing physical relations, the selection of the reconstructed variables, the required unit conversions, and the normalization applied to the model outputs.

Combining the previous equations, the complete mapping is written as
\begin{equation}
\widehat{\mathbf{x}}_t
=
\mathcal{P}\!\left(
    \mathbf{w}_t,
    \mathbf{p}_{0}\odot
    \mathcal{R}\!\left(
        \mathcal{E}_{\phi}\!\left(\mathbf{x}_t\right)
    \right);
    \boldsymbol{\Theta}
\right).
\label{eq:complete_physics_ae}
\end{equation}

For a training set containing $N$ observations, the encoder parameters can be estimated by minimizing a composite loss. The reconstruction term is expressed through Equation \ref{eq:reconstruction_loss}:
\begin{equation}
\mathcal{L}_{\mathrm{rec}}(\phi)
=
\frac{1}{N}
\sum_{k=1}^{N}
\left\|
\mathbf{x}_{t_k}
-
\widehat{\mathbf{x}}_{t_k}
\right\|_2^2.
\label{eq:reconstruction_loss}
\end{equation}
On top of this term, additional terms can introduce more physical knowledge in the form of learning bias. The general objective can be formulated as
\begin{equation}
\mathcal{L}(\phi)
=
\mathcal{L}_{\mathrm{rec}}(\phi)
+
\sum_{j=1}^{J}
\lambda_j\mathcal{L}_{j}(\phi),
\label{eq:composite_loss}
\end{equation}
where $\mathcal{L}_{j}$ represents an application-specific physical relation or constraint and $\lambda_j$ controls its contribution. These terms introduce a learning bias complementary to the inductive bias imposed by the equation-based decoder.

In this formulation, the physical decoder contains no trainable neural layers. Huang et al.~\cite{Huang2024PhysicsInformedAE} introduced a related architecture in which the physical equations were followed by conventional trainable layers within the decoder. In the present formulation, the normalized outputs of the physical equations directly constitute the reconstructed measurement vector. The resulting structure requires a sufficiently detailed and differentiable description of the monitored process, since the physical model must account for the complete mapping from the inferred component parameters to the reconstructed variables. This restriction also prevents additional decoder layers from compensating for inaccuracies in the inferred physical modifiers or from encoding degradation-related information through trainable parameters.

Training is consequently restricted to identifying the encoder parameters $\phi$ that generate physically admissible modifiers capable of reconstructing the observed system response through $\mathcal{P}(\cdot)$. The imposed model structure specifies the physical quantity associated with each latent coordinate and the equations through which changes in that quantity influence the measurements. The architecture therefore provides a structural interpretation of the latent representation and enables the individual latent trajectories to be examined at the level of components or degradation mechanisms.

\subsection{Operating-Condition Correction of the Latent Representation}
\label{subsec:operating_condition_correction}

The physical interpretation assigned to the latent coordinates does not ensure that their values depend exclusively on degradation. In systems operating under variable conditions, the physical quantities represented in the latent space may change across the operating envelope even when the system remains healthy (e.g., turbofan engines \cite{wen2024residual}). The $j$-th latent variable can therefore be decomposed as
\begin{equation}
z_{j,t}
=
z_j^{\mathrm{H}}\!\left(\mathbf{w}_t\right)
+
\delta z_{j,t}^{\mathrm{D}}
+
\varepsilon_{j,t},
\label{eq:general_latent_decomposition}
\end{equation}

where $z_j^{\mathrm{H}}\!\left(\mathbf{w}_t\right)$ denotes the value expected for the $j$-th latent quantity under healthy conditions at operating point $\mathbf{w}_t$,
$\delta z_{j,t}^{\mathrm{D}}$ denotes the deviation associated with degradation, and
$\varepsilon_{j,t}$ accounts for estimation errors, model mismatch, and unmodelled effects.
According to Eq.~\eqref{eq:general_latent_decomposition}, variations observed in the raw latent trajectories may result from changes in the operating conditions, degradation, or their combined effect. A direct use of $\mathbf{z}_t$ for health assessment may therefore transfer operating-regime variability into the resulting HI. The healthy operating-condition contribution must be estimated whenever its magnitude is not negligible with respect to the degradation-related variation.

Let $\mathcal{H}$ denote a reference set containing observations representative of healthy system operation:
\begin{equation}
\mathcal{H}
=
\left\{
\left(
\mathbf{w}_t,
\mathbf{z}_t
\right)
:
t\in\mathcal{T}_{\mathrm{H}}
\right\},
\label{eq:healthy_reference_dataset}
\end{equation}

where $\mathcal{T}_{\mathrm{H}}$ is the set of time indices associated with the healthy reference condition.
For each latent coordinate $j\in\{1,\ldots,n_z\}$, a model
$h_j:\mathbb{R}^{n_w}\rightarrow\mathbb{R}$ is identified from the healthy
reference set to estimate its dependence on the operating conditions:
\begin{equation}
\widehat{h}_j
=
\underset{h_j}{\arg\min}
\sum_{t\in\mathcal{T}_{\mathrm{H}}}
\left[
z_{j,t}
-
h_j\!\left(\mathbf{w}_t\right)
\right]^2.
\label{eq:general_healthy_operating_model}
\end{equation}
The fitted function $\widehat{h}_j(\mathbf{w}_t)$ estimates the value expected
for the $j$-th latent variable under healthy conditions at the operating point
$\mathbf{w}_t$. The corrected latent coordinate is then computed as
$\widetilde{z}_{j,t}
=
z_{j,t}
-
\widehat{z}_j^{\mathrm{H}}\!\left(\mathbf{w}_t\right).$
and the vector of corrected latent variables is consequently defined as
\begin{equation}
\widetilde{\mathbf{z}}_t
=
\mathbf{z}_t
-
\widehat{\mathbf{z}}^{\mathrm{H}}\!\left(\mathbf{w}_t\right),
\qquad
\widehat{\mathbf{z}}^{\mathrm{H}}\!\left(\mathbf{w}_t\right)
=
\begin{bmatrix}
\widehat{z}_1^{\mathrm{H}}\!\left(\mathbf{w}_t\right) &
\cdots &
\widehat{z}_{n_z}^{\mathrm{H}}\!\left(\mathbf{w}_t\right)
\end{bmatrix}^{\top}.
\label{eq:corrected_healthy_latent_vectors}
\end{equation}

Under an adequate representation of the healthy operating-condition dependence, the corrected coordinate $\widetilde{z}_{j,t}$ approximates the degradation-related contribution:
$\widetilde{z}_{j,t}
\approx
\delta z_{j,t}^{\mathrm{D}}
+
\varepsilon_{j,t}.$

The residual representation $\widetilde{\mathbf{z}}_t$ can be retained as a multidimensional description of the health state, with individual coordinates associated with specific components or degradation mechanisms. It can also be combined into a scalar HI through a subsequent aggregation function
$\mathcal{G}:\mathbb{R}^{n_z}\rightarrow\mathbb{R}$:
$\mathrm{HI}_t
=
\mathcal{G}\!\left(
\widetilde{\mathbf{z}}_t
\right).$
The selection of the healthy reference set, the function class
$\mathcal{F}_j$, and the aggregation function $\mathcal{G}(\cdot)$ depends on the available knowledge and on the characteristics of the considered application. Their specific implementation for the turbofan case study is described in Section~\ref{sec:case_study}.
For component-level comparisons, the corrected latent contribution is added to the healthy baseline evaluated at a common reference operating condition,
\begin{equation}
\mathbf{z}_{t,\mathrm{ref}}
=
\hat{\mathbf{z}}^{\mathrm{H}}(\mathbf{w}_{\mathrm{ref}})
+
\tilde{\mathbf{z}}_t.
\label{eq:reference_condition_latent}
\end{equation}
This operation retains the inferred degradation-related contribution while replacing the healthy operating-condition contribution with its value at $\mathbf{w}_{\mathrm{ref}}$. The resulting latent coordinates are then converted into physical quantities through the decoder mappings, allowing their evolution to be compared at a common operating point.

\section{Case Study: Physics-Decoder Autoencoder for N-CMAPSS}
\label{sec:case_study}

\begin{figure*}
\centering
\includegraphics[width=0.95\textwidth]{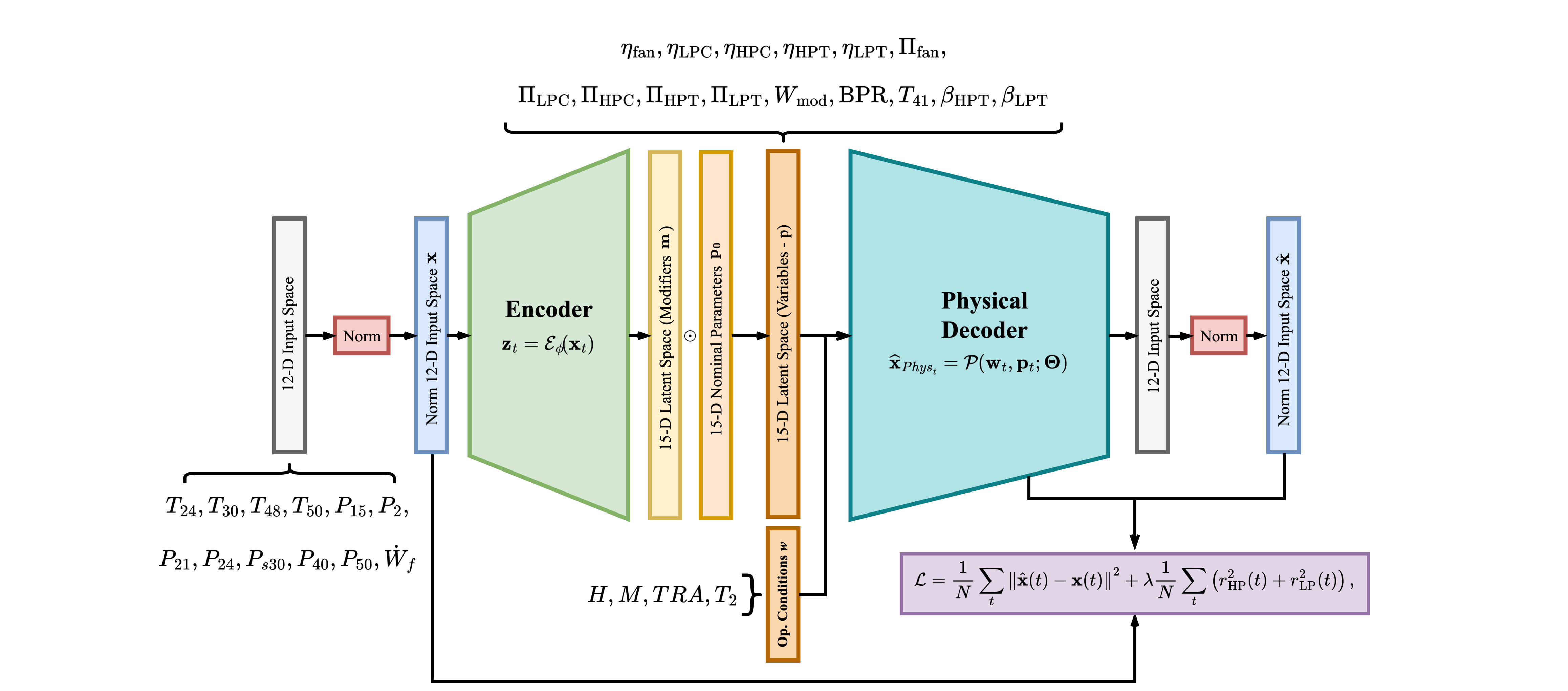}
\caption{Implementation of the physics-constrained autoencoder for the N-CMAPSS turbofan application. The $15$-dimensional latent space represents physically assigned engine parameters, which are combined with the operating conditions in the physical decoder. Model training minimizes the reconstruction error together with the residuals associated with the turbine shaft-power balances.}
\label{fig:ncm_model}
\end{figure*}

This section specializes the general framework to a turbofan-engine case study based on the N-CMAPSS dataset. The objective is to show how a physics-structured decoder can be implemented when the measured variables are thermodynamic and mechanical quantities associated with engine operation \cite{abed2024implementing}. The specific implementation of the physics constrained autoencoder for the N-CMAPSS turbofan application is reported in Figure \ref{fig:ncm_model}.

\subsection{N-CMAPSS Dataset}
\label{subsec:ncmapss_dataset}
 
The New Commercial Modular Aero-Propulsion System Simulation (N-CMAPSS) dataset provides run-to-failure trajectories for fleets of simulated turbofan engines operating under realistic flight conditions \cite{AriasChao2021NCMAPSS}. It is suitable for the present study because, on top of being one of the most popular datasets of PHM evaluation, it contains operating conditions, multivariate sensor measurements, and degradation trajectories generated using a physics-based turbofan simulation environment. Among its eight subsets, DS003 is selected as the main case study for model development, HI construction, and benchmarking. It represents a combined degradation mode affecting the efficiency and flow capacity of the low-pressure turbine and the efficiency of the high-pressure turbine. DS004--DS006, which introduce degradation in the fan and compressor sections, are subsequently used to test the same architecture across different degradation modes and assess the component-level diagnostic information retained by a model trained on the pooled scenarios. For each DS003 engine, the dataset contains 14 sensor variables collected throughout its operating life, from the initial healthy condition to failure. Twelve variables, denoted by $\mathbf{X}$, are retained in the present formulation. The fan and core shaft speeds, $N_f$ and $N_c$, are excluded from the reconstruction targets because, as customarily done in simple turbofan modeling, the adopted thermodynamic formulation does not explicitly relate spool speeds to the reconstructed gas-path quantities \cite{walsh2004gas}. The corresponding operating conditions, denoted by $\mathbf{W}$, comprise altitude, Mach number, throttle-resolver angle, and total temperature at the fan inlet. The dataset also provides the model health parameters, $\boldsymbol{\theta}$, comprising efficiency and flow-capacity modifiers for the fan, Low-Pressure Compressor (LPC), High-Pressure Compressor (HPC), High-Pressure Turbine (HPT), and Low-Pressure Turbine (LPT). These parameters describe the component degradation imposed during data generation and provide reference trajectories for validating the inferred latent quantities. The ground-truth HI, $h_{\mathrm{GT}}$, is obtained from operational margins evaluated under reference conditions, with failure reached when $h_{\mathrm{GT}}$ decreases to zero.

\begin{figure*}
    \centering
    \includegraphics[width=\textwidth]{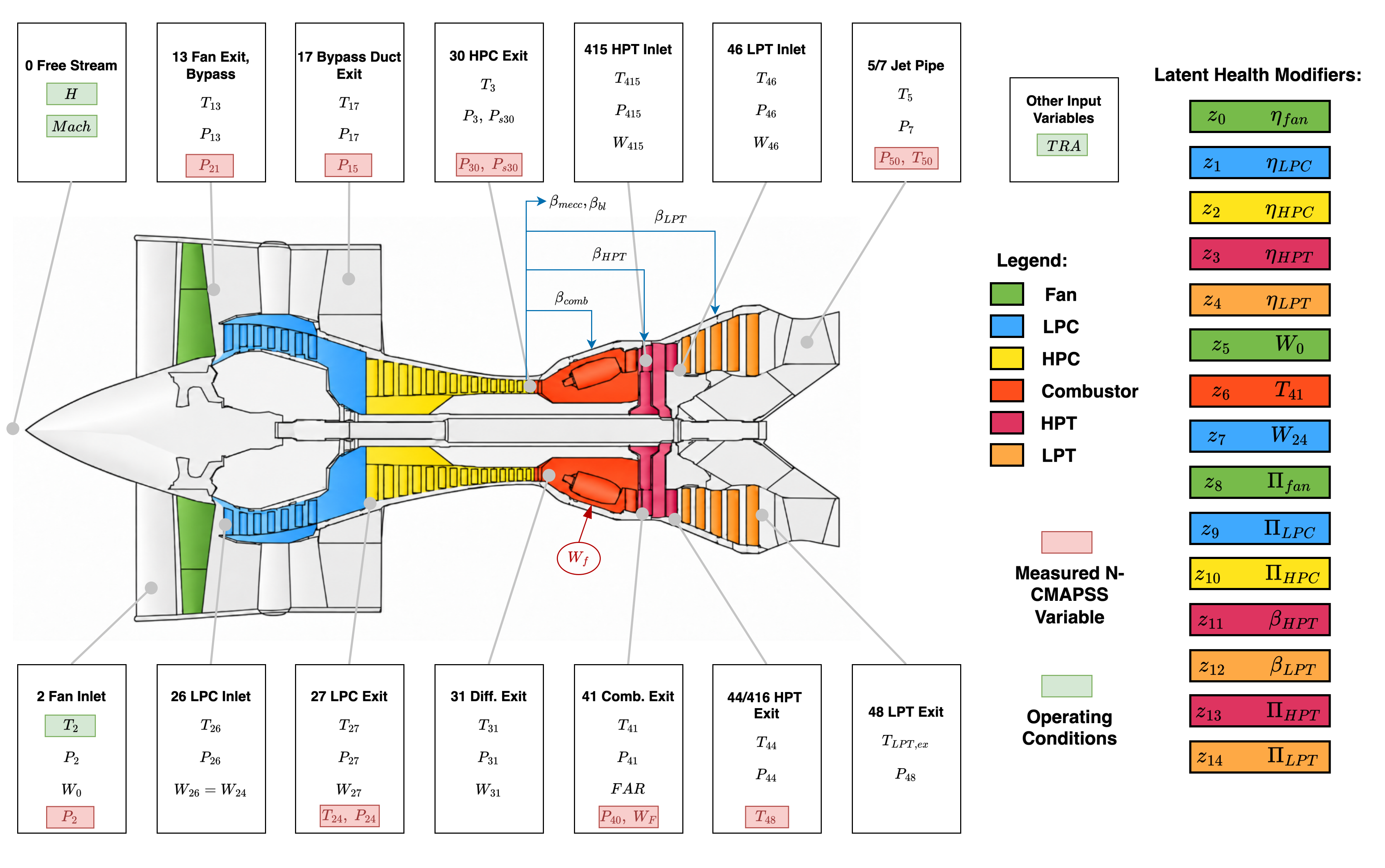}
    \caption{Schematic of the physics-based turbofan decoder and its correspondence with the N-CMAPSS variables and latent health modifiers. The engine stations and component decomposition are shown together with the measured N-CMAPSS variables used as model inputs and reconstruction targets, highlighted in red. Component colors identify the fan, LPC, HPC, compressor, HPT, and LPT. The latent variables $\mathbf{z}$ parameterize component-level health modifiers associated with efficiencies, pressure ratios, flow capacity, and selected thermodynamic quantities, while the operating conditions enter through the free-stream variables.}
    \label{fig:physics_decoder_turbofan}
\end{figure*}

\subsection{Equations}
\label{subsec:physical_modifiers}
\subsubsection{Differentiable equation-based decoder}
\label{sec:physics_decoder}

The physical decoder implements a simplified two-spool turbofan model as a differentiable computation graph, using textbook thermodynamic equations from \cite{walsh2004gas}. Internal thermodynamic states follow conventional engine station numbering, while measured and reconstructed channels retain their N-CMAPSS names to facilitate comparison with the dataset. For clarity, Figure~\ref{fig:physics_decoder_turbofan} brings these conventions together: it shows the engine stations, the N-CMAPSS channels in red boxes, the cooling flows, and the latent health modifiers associated with each component. The correspondence between station labels and N-CMAPSS variables is also listed in Appendix~\ref{app:NCMAPSStable}, and Table~\ref{tab:decoder_symbols} summarizes the symbols used throughout the formulation. The decoder receives altitude $H(t)$, flight Mach number $M(t)$, throttle-resolver angle $\mathrm{TRA}(t)$, and fan-inlet total temperature $T_2(t)$ as operating conditions. Each latent variable adjusts a nominal design quantity through a positive multiplicative modifier:
\begin{equation}
q_k(t)=q_k^{0}\,\mathrm{mod}_{k,t},
\qquad
\mathrm{mod}_{k,t}=r\!\left(z_{k,t}\right)=e^{0.05\,z_{k,t}},
\label{eq:modifier}
\end{equation}
The exponential mapping is smooth and strictly increasing, preserves positivity, and gives $\mathrm{mod}_{k,t}=1$ at $z_{k,t}=0$, recovering the calibrated nominal engine at the latent origin. It permits both increases and decreases relative to the nominal value without imposing a fixed range on the modifier. The coefficient $0.05$ sets the latent-to-modifier scale: near the origin, $\mathrm{mod}_{k,t}\approx1+0.05\,z_{k,t}$, so a unit latent change corresponds approximately to a $5\%$ relative change in the associated quantity.The set of equations that we will present in the following paragraphs requires 16 quantities as input. These quantities are:
\begin{equation}
\begin{split}
\mathbf{p}(t)=\{&
\eta_{\mathrm{fan}},\eta_{\mathrm{LPC}},\eta_{\mathrm{HPC}},
\eta_{\mathrm{HPT}},\eta_{\mathrm{LPT}},
\Pi_{\mathrm{fan}},\Pi_{\mathrm{LPC}},\Pi_{\mathrm{HPC}},
\Pi_{\mathrm{HPT}},\Pi_{\mathrm{LPT}},\\
&W_{\mathrm{mod}},\mathrm{BPR},T_{41},\beta_{\mathrm{comb}},
\beta_{\mathrm{HPT}},\beta_{\mathrm{LPT}}
\},
\end{split}
\label{eq:turbofan_parameters}
\end{equation}
comprising five component efficiencies, three compressor pressure ratios, the two
turbine expansion ratios, the inlet flow capacity, the bypass ratio, the
turbine-inlet temperature level, and three cooling-flow fractions.


\begin{table*}[t]
\centering
\caption{Symbols and physical quantities used in the turbofan decoder.}
\label{tab:decoder_symbols}
\renewcommand{\arraystretch}{0.96}
\setlength{\tabcolsep}{2.5pt}
\scriptsize
\begin{tabularx}{\textwidth}{p{1.55cm} lX lX lX}
\toprule
\textbf{Category} &
\textbf{Symbol} & \textbf{Meaning} &
\textbf{Symbol} & \textbf{Meaning} &
\textbf{Symbol} & \textbf{Meaning} \\
\midrule

Operating
conditions
&
$H$ & Altitude &
$M$ & Flight Mach number &
$T_2$ & Fan-inlet total temperature \\

&
$\mathrm{TRA}$ & Throttle-resolver angle &
$\widehat{\mathrm{TRA}}$ & Normalised throttle setting &
& \\
\midrule

Atmospheric
quantities
&
$T_{\mathrm{amb}}$ & Ambient static temperature &
$P_{\mathrm{amb}}$ & Ambient static pressure &
$P_{\mathrm{SL}}$ & Sea-level pressure \\

&
$P_0$ & Freestream total pressure &
$k_H$ & ISA altitude coefficient &
$R$ & Specific gas constant \\
\midrule

Gas
properties
&
$\gamma$ & Specific-heat ratio, air &
$\gamma_t$ & Specific-heat ratio, turbine gas &
$c_p$ & Specific heat, air \\

&
$c_{p,t}$ & Specific heat, turbine gas &
$c_{p,\mathrm{mix}}$ & Mixed-stream specific heat &
$\operatorname{MFP}$ & Mass-flow parameter \\

&
$\mathrm{LHV}$ & Fuel lower heating value &
$\eta_{\mathrm{comb}}$ & Combustion efficiency &
$\eta_m$ & Mechanical efficiency \\
\midrule

Component
efficiencies
&
$\eta_{\mathrm{fan}}$ & Fan efficiency &
$\eta_{\mathrm{LPC}}$ & LPC efficiency &
$\eta_{\mathrm{HPC}}$ & HPC efficiency \\

&
$\eta_{\mathrm{HPT}}$ & HPT efficiency &
$\eta_{\mathrm{LPT}}$ & LPT efficiency &
$\eta_c,\eta_t$ & Generic compressor, turbine efficiency \\
\midrule

Pressure
ratios
&
$\Pi_{\mathrm{fan}}$ & Fan pressure ratio &
$\Pi_{\mathrm{LPC}}$ & LPC pressure ratio &
$\Pi_{\mathrm{HPC}}$ & HPC pressure ratio \\

&
$\Pi_{\mathrm{HPT}}$ & HPT expansion ratio &
$\Pi_{\mathrm{LPT}}$ & LPT expansion ratio &
$\Pi_c,\Pi_t$ & Generic compressor, turbine ratio \\

&
$\delta$ & Pressure-loss fraction &
$\delta_{\mathrm{in}}$ & Inlet pressure-loss fraction &
$P_{\mathrm{in}},P_{\mathrm{out}}$ & Element inlet, outlet pressure \\
\midrule

Mass-flow
quantities
&
$W_0$ & Total inlet mass flow &
$W_{\mathrm{core}}$ & Core mass flow &
$W_{\mathrm{bypass}}$ & Bypass mass flow \\

&
$W_3$ & HPC-exit mass flow &
$W_{31}$ & Combustor-inlet airflow &
$W_{\mathrm{comb}}$ & Airflow available for combustion \\

&
$W_{415},W_{46}$ & HPT, LPT gas-path mass flow &
$W_f$ & Fuel mass flow &
$W_{\mathrm{mod}}$ & Inlet flow-capacity modifier \\

&
$\beta_{\mathrm{bl}}$ & Customer-bleed fraction &
$\beta_{\mathrm{mec}}$ & Mechanical-offtake fraction &
$A_{\mathrm{in}}$ & Effective inlet area \\
\midrule

Flow
distribution
&
$\mathrm{BPR}$ & Bypass ratio &

$\beta_{\mathrm{comb}}$ & Combustor-cooling fraction \\

&
$\beta_{\mathrm{HPT}}$ & HPT-cooling fraction &
$\beta_{\mathrm{LPT}}$ & LPT-cooling fraction &
& \\
\midrule

Thermal
quantities
&
$T_{\mathrm{in}},T_{\mathrm{out}}$ & Element inlet, outlet temperature &
$T_{\mathrm{mix}}$ & Mixed-stream temperature &
$T_{41}$ & Turbine-inlet temperature (SOT) \\

&
$T_{41}^{0}$ & Idle turbine-inlet temperature &
$T_{41}'$ & Throttle sensitivity of $T_{41}$ &
$P_i,T_i$ & Conditions at station $i$ \\

&
$T_{\mathrm{LPT,ex}}$ & LPT exit temperature, before exit mixing & \\
\midrule

Power
balance
&
$\dot{Q}_{\mathrm{fan}}$ & Fan power &
$\dot{Q}_{\mathrm{LPC}}$ & LPC power &
$\dot{Q}_{\mathrm{HPC}}$ & HPC power \\

&
$r_{\mathrm{HP}}$ & HP-shaft power residual &
$r_{\mathrm{LP}}$ & LP-shaft power residual &
$\lambda$ & Power-balance penalty weight \\
\midrule

Output
&
$\mathbf{x}_{\mathrm{phys}}$ & Decoder output vector &
$\hat{\mathbf{x}}$ & Normalised reconstruction &
$\mathcal{L}$ & Training objective \\

\bottomrule
\end{tabularx}
\end{table*}

The inlet conditions are calculated using an International Standard Atmosphere
approximation and the isentropic ram relations:
\begin{align}
P_{\mathrm{amb}}
&=
P_{\mathrm{SL}}
\left(1-k_H H\right)^{5.25588},
&
T_{\mathrm{amb}}
&=
\frac{T_2}
{1+\frac{\gamma-1}{2}M^2},
&
P_{0}
&=
P_{\mathrm{amb}}
\left(\frac{T_2}{T_{\mathrm{amb}}}\right)^{\frac{\gamma}{\gamma-1}},
&
P_2
&=
(1-\delta_{\mathrm{in}})P_0.
\label{eq:inlet_conditions}
\end{align}

The inlet mass flow is obtained from the compressible mass-flow parameter:
\begin{align}
\operatorname{MFP}(M)
&=
\frac{
M\sqrt{\gamma/R}
}{
\left(1+\frac{\gamma-1}{2}M^2\right)^{
\frac{\gamma+1}{2(\gamma-1)}
}},
&
W_0
&=
mod_{W_0}
\frac{P_2 A_{\mathrm{in}}\operatorname{MFP}(M)}
{\sqrt{T_2}}.
\label{eq:inlet_mass_flow}
\end{align}

Pressure losses across the inlet, ducts, diffuser, combustor, and jet pipe are
represented as
\begin{equation}
P_{\mathrm{out}}
=
\left(1-\delta\right)P_{\mathrm{in}}.
\label{eq:pressure_loss}
\end{equation}

The fan and compressors are described by
\begin{align}
P_{\mathrm{out}}
&=
\Pi_c P_{\mathrm{in}},
\label{eq:compressor_pressure}
\\
T_{\mathrm{out}}
&=
T_{\mathrm{in}}
\left[
1+
\frac{
\Pi_c^{(\gamma-1)/\gamma}-1
}{
\eta_c
}
\right].
\label{eq:compressor_temperature}
\end{align}
Equations~\eqref{eq:compressor_pressure} and~\eqref{eq:compressor_temperature}
are applied sequentially to the fan, low-pressure compressor, and high-pressure
compressor using the corresponding latent-modified efficiencies and pressure
ratios.

The HPC exit requires one distinction that the remaining stations do not. The
decoder propagates total pressures throughout, whereas the N-CMAPSS channel
$P_{s30}$ is a static pressure and $P_{41}$  a total one. Both are
reported from the single computed quantity $P_3$, so the ratio between them is
fixed by the decoder and it must equal the ratio the dataset holds. To further examine this behaviour, the two ratios were evaluated on the healthy development split and found to remain constant across the operating conditions represented in N-CMAPSS.
\begin{equation}
\frac{P_{30}}{P_{s30}} = 1.0660 \;\; (\mathrm{CV}\; 0.21\,\%),
\qquad
\frac{P_{41}}{P_{30}} = 0.9550 \;\; (\mathrm{CV}\; <10^{-4}\,\%),
\label{eq:ps30_ratios}
\end{equation}
the first being the dynamic-head contribution at the compressor exit and the
second the combined diffuser and combustor total-pressure loss. The two reported
pressures are therefore formed as
\begin{equation}
P_{s30} = \frac{P_3}{1.0660},
\qquad
P_{41} = 0.9550\, P_3,
\label{eq:hpc_exit_pressures}
\end{equation}
while $P_{31}=(1-\delta_{\mathrm{diff}})P_3$.

The fan flow is divided into core and bypass streams according to
\begin{equation}
W_{\mathrm{core}}
=
\frac{W_0}{1+\mathrm{BPR}},
\qquad
W_{\mathrm{bypass}}
=
W_0-W_{\mathrm{core}}.
\label{eq:flow_split}
\end{equation}
Customer bleed and mechanical offtakes are extracted at the HPC as prescribed
fractions of the core flow, $W_3=W_{\mathrm{core}}
(1-\beta_{\mathrm{bl}}-\beta_{\mathrm{mec}})$. 
The cooling flows are then extracted at the compressor delivery, all of them as fractions of the core flow. The combustor cooling rejoins in the combustor, and the two turbine cooling streams bypass the combustor: the HPT stream rejoins the gas path at the HPT inlet, where it mixes with the combustor exhaust and expands through the high-pressure turbine, whereas the LPT stream rejoins downstream of the low-pressure turbine, in the jet pipe. Over the $425{,}156$ test samples the resulting core mass imbalance is around $-0.5\%$, which is the mechanical offtake $\beta_{\mathrm{mec}}$ leaving the engine.:
\begin{equation}
W_{31}
=
W_{3}
-
W_{\mathrm{core}}
\left(\beta_{\mathrm{comb}}
+\beta_{\mathrm{HPT}}+\beta_{\mathrm{LPT}}\right),
\qquad
W_{\mathrm{comb}}
=
W_{31}+W_{\mathrm{core}}\beta_{\mathrm{comb}}
=
W_{3}-W_{\mathrm{core}}
\left(\beta_{\mathrm{HPT}}+\beta_{\mathrm{LPT}}\right).
\label{eq:combustor_airflow}
\end{equation}

The turbine-inlet temperature is set by the throttle demand and scaled by its
latent modifier, and the fuel flow follows from the combustor energy balance:
\begin{equation}
T_{41}
=
\left(T_{41}^{0}+T_{41}'\,\widehat{\mathrm{TRA}}\right)mod_{T_{41}},
\qquad
W_{f}
=
W_{\mathrm{comb}}
\frac{c_{p,t}T_{41}-c_pT_{31}}
{\eta_{\mathrm{comb}}\mathrm{LHV}-c_{p,t}T_{41}},
\label{eq:combustor}
\end{equation}
with $\widehat{\mathrm{TRA}}=\mathrm{TRA}/100\in[0,1]$. Cooling flows are mixed
with the main gas path through the enthalpy balance
\begin{equation}
T_{\mathrm{mix}}
=
\frac{
\sum_j W_j c_{p,j}T_j
}{
c_{p,t}\sum_jW_j
},
\label{eq:mixing}
\end{equation}
applied at the HPT inlet, where the HPT cooling air rejoins the combustor exhaust, and at the jet pipe, where the LPT cooling air and the customer bleed rejoin the core stream.

For the turbines, the expansion ratio of each stage is itself a latent modifier
and the efficiency sets the temperature drop through the isentropic expansion
relation:
\begin{align}
\Pi_t
&=
\Pi_t^{0}\,e^{0.05 z},
&
T_{\mathrm{out}}
&=
T_{\mathrm{in}}
\left[
1-\eta_t
\left(
1-\Pi_t^{\frac{1-\gamma_t}{\gamma_t}}
\right)
\right],
&
P_{\mathrm{out}}
&=
\frac{P_{\mathrm{in}}}{\Pi_t}.
\label{eq:turbine_expansion}
\end{align}

The combustor exit pressure follows Eq. \ref{eq:hpc_exit_pressures} and, downstream of the HPT, temperature and pressure are identified by
$T_{44}$, $P_{44}$, with then $T_{46} = T_{44}$ and $P_{46} = (1-\delta_{\rm itd})P_{44}$ at the LPT inlet after the inter-turbine duct. Applied to the low-pressure turbine, Equation \ref{eq:turbine_expansion} gives the LPT exit state $(T_{\mathrm{LPT,ex}}, P_{48})$: the channel T50 is the mixed exit temperature $T_5$ of Equation \ref{eq:mixing}, formed once the LPT cooling air and the customer bleed have rejoined the core stream, and the channel P50 is the jet-pipe pressure $P_7$ obtained from $P_{48}$ through Equation \ref{eq:pressure_loss}. The jet-pipe temperature follows from
Eq.~\eqref{eq:mixing} applied to the LPT cooling air and the customer bleed
rejoining the core.
The shaft power balance becomes a soft constraint. The required powers are
\begin{equation}
\dot{Q}_{\mathrm{fan}}
=
W_0c_p(T_{13}-T_2),
\quad
\dot{Q}_{\mathrm{LPC}}
=
W_{\mathrm{core}}c_p(T_{27}-T_{26}),
\quad
\dot{Q}_{\mathrm{HPC}}
=
W_{3}c_p(T_3-T_{27}),
\label{eq:shaft_power}
\end{equation}
and the relative imbalance on each shaft is
\begin{align}
r_{\mathrm{HP}}
&=
\frac{
\eta_mW_{415}c_{p,t}\left(T_{415}-T_{44}\right)
-\dot{Q}_{\mathrm{HPC}}
}{
\left|\dot{Q}_{\mathrm{HPC}}\right|
},
&
r_{\mathrm{LP}}
&=
\frac{
\eta_mW_{46}c_{p,t}\left(T_{46}-T_{\mathrm{LPT,ex}}\right)
-\left(\dot{Q}_{\mathrm{fan}}+\dot{Q}_{\mathrm{LPC}}\right)
}{
\left|\dot{Q}_{\mathrm{fan}}+\dot{Q}_{\mathrm{LPC}}\right|
}
\label{eq:power_residual}
\end{align}

The equations are evaluated sequentially along the inlet, fan, bypass duct, LPC,
HPC, combustor, HPT, LPT, and exhaust paths. The physical-decoder output is assembled as

\begin{equation}
\widehat{\mathbf{x}}_{\mathrm{phys}}(t)
=
\begin{bmatrix}
T_{24} \quad T_{30} \quad T_{48} \quad T_{50} \quad P_{15} \quad P_{2} \quad P_{21} \quad P_{24} \quad P_{s30} \quad P_{40} \quad P_{50} \quad W_f
\end{bmatrix}^{\top}.
\label{eq:decoder_output}
\end{equation}

corresponding to the selected N-CMAPSS variables. The outputs are finally
converted to the N-CMAPSS measurement units and normalized using the bounds
adopted for model training, giving $\hat{\mathbf{x}}(t)$. As already mentioned, the shaft-speed variables $N_f$ and $N_c$ were excluded because the present decoder does not explicitly model spool-speed dynamics. The network is trained
on the reconstruction error with the two power residuals as penalty terms, thus representing a form of learning bias via the residual term
\begin{equation}
\mathcal{L}
=
\frac{1}{N}\sum_t
\left\|
\hat{\mathbf{x}}(t)-\mathbf{x}(t)
\right\|^2
+
\lambda
\frac{1}{N}\sum_t
\left(
r_{\mathrm{HP}}^2(t)+r_{\mathrm{LP}}^2(t)
\right),
\qquad
\lambda=1,
\label{eq:objective}
\end{equation}
so that the shaft balance is satisfied where it is compatible with the
measurements instead of being imposed regardless of them. At the trained solution
the constraint is nonetheless close to being met: over the $425{,}156$ test
samples the mean relative imbalance is around $0.111\%$ on the HP shaft and around $0.119\%$ on the LP shaft, the $95$th percentile is below $0.35\%$. This small residual allows the two turbine efficiencies to vary independently within the decoder while maintaining approximate shaft balance. Their separate recovery from the measurements remains subject to the identifiability analysis in Section~\ref{subsec:latent_identifiability}.

\subsubsection{Latent-space sensitivity and dimensionality selection}
As reported in Eq. \ref{eq:turbofan_parameters}, the physical decoder initially defines a 16-dimensional latent space, with each
coordinate associated with one of the physical quantities entering the model.
However, the availability of twelve measured variables does not guarantee that
all 16 latent quantities are observable from the reconstruction problem. Sensitivity measures the influence of a coordinate on the outputs.
The sensitivity of the reconstructed measurements to each latent coordinate was
therefore evaluated through the Jacobian
$\partial\hat{\mathbf{x}}/\partial z_k$. 
The sensitivity analysis hence guides the selection of the latent coordinates, with methodological details provided in Appendix~\ref{app:observability}.


In particular, one coordinate has a markedly lower sensitivity than the remaining latent
variables. The Jacobian norm associated with $\beta_{\mathrm{comb}}$ is in fact $3.77\times10^{-5}$. For comparison,
the lowest value among the retained coordinates is
$1.80\times10^{-3}$, corresponding to $\beta_{\mathrm{LPT}}$.
The combustor cooling fraction $\beta_{\mathrm{comb}}$ is consequently fixed at its design value, reducing the latent space from 16 to 15 dimensions.
The complete sensitivity comparison and the ablation obtained by reintroducing this coordinate are reported in Appendix~\ref{app:observability}.

The final latent representation adopted in the proposed model is therefore
\[
\mathbf{p}(t) =
\{\eta_{\mathrm{fan}}, \eta_{\mathrm{LPC}}, \eta_{\mathrm{HPC}},
  \eta_{\mathrm{HPT}}, \eta_{\mathrm{LPT}},
  \Pi_{\mathrm{fan}}, \Pi_{\mathrm{LPC}}, \Pi_{\mathrm{HPC}},
  \Pi_{\mathrm{HPT}}, \Pi_{\mathrm{LPT}},
  W_{\mathrm{mod}}, \mathrm{BPR}, T_{41},
  \beta_{\mathrm{HPT}}, \beta_{\mathrm{LPT}}\},
\]
comprising five component efficiencies, three compressor pressure ratios, two
turbine expansion ratios, the inlet flow capacity, the bypass ratio, the
turbine-inlet temperature, and two turbine cooling fractions. An overview of
the resulting architecture, including the measured inputs and the selected
15 latent variables, is provided in Figure~\ref{fig:ncm_model}.

With 12 measured variables and 15 latent coordinates, the resulting architecture
is formally over-complete. In this case, however, the decoder contains no
trainable parameters and the latent variables act exclusively through the fixed
thermodynamic relations. The wider latent space therefore cannot be exploited by the decoder to learn an unconstrained identity mapping.

\subsection{Model Implementation}
\label{subsec:model_implementation}

The model was implemented in Python using TensorFlow~2.10 and Keras with 32-bit floating-point precision. The measured variables were min--max normalized using bounds estimated from the training set, while the operating conditions were provided to the decoder in their original units.
The encoder comprises five fully connected layers with $256$ neurons and ReLU activations. 
The decoder is a fixed differentiable computation graph implementing the thermodynamic relations introduced in Section~\ref{sec:physics_decoder}. It receives the latent variables together with altitude, Mach number, throttle-resolver angle, and fan-inlet total temperature, and contains no trainable parameters.
All experiments were executed on identically configured nodes of a SLURM cluster equipped with two Intel Xeon E5-2698 v4 CPUs at $2.20$~GHz, $450$~GB of RAM, and eight NVIDIA Tesla V100-SXM2 GPUs with $32$~GB of memory. Each job was allocated one GPU, four CPU cores, and $32$~GB of RAM. The median training time was 9.4 min per model on a single V100 GPU over five runs, ranging from 8.9 to 11.5 min due to early stopping.
Each model was trained with the Adam optimizer, a learning rate of $10^{-3}$, a
batch size of 512, gradient-norm clipping at 1.0, and an epoch cap of 120 with early stopping on the validation loss (patience 20, best weights restored). The cap is never reached: across the five seeds training stops between epoch 26 and 52 and the best validation loss is reached between epoch 6 and 32. 


\section{Results} \label{sec:results} 

\begin{figure*}
\centering
\includegraphics[width=0.95\textwidth]{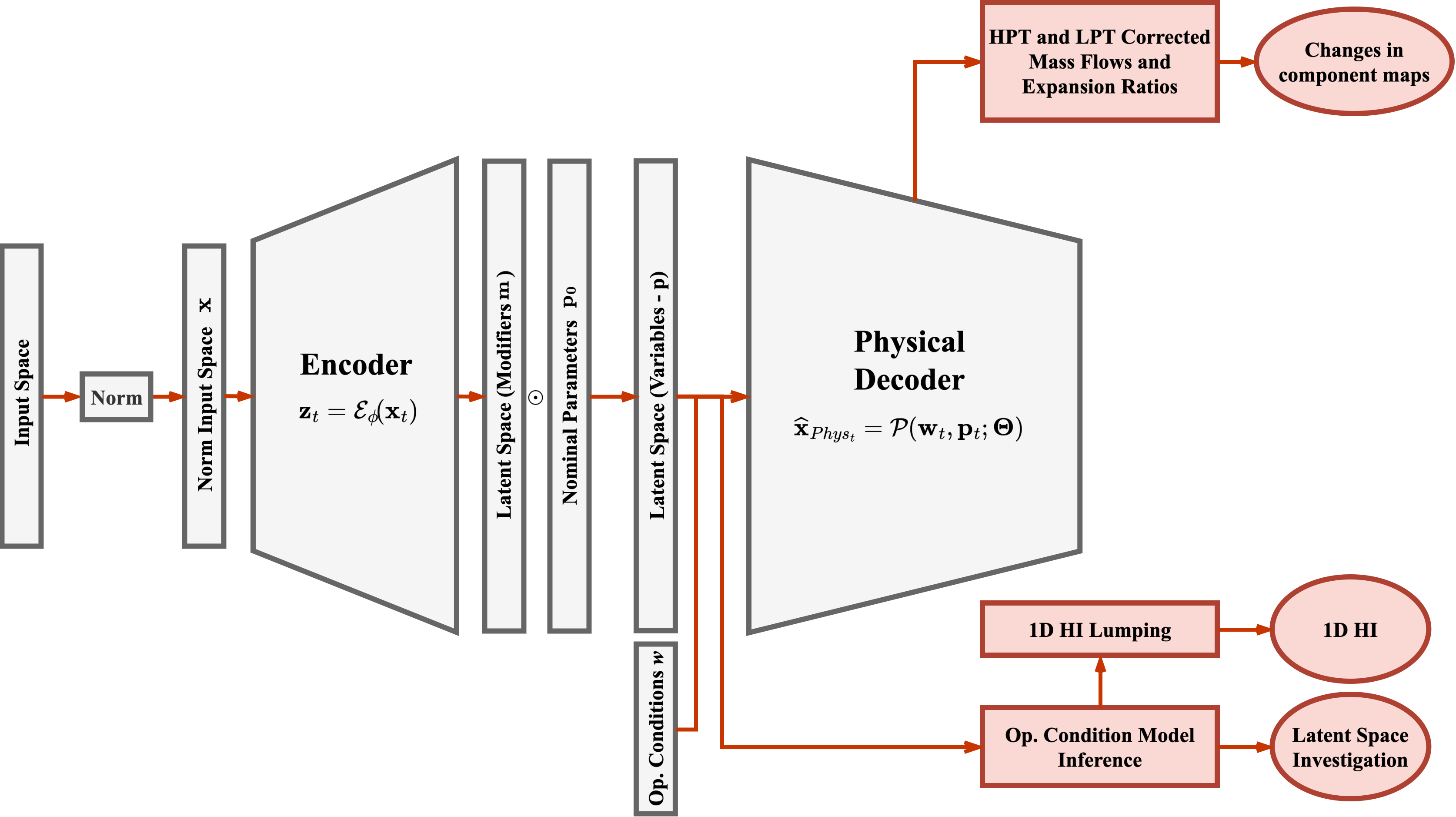}
\caption{Inference and health-assessment workflow of the proposed framework. The inferred physical latent variables are used to investigate the latent space and construct the system-level HI after removal of the operating-condition contribution. In parallel, the inferred turbine quantities are used to evaluate degradation-induced changes in the component operating maps.}
\label{fig:ncm_inference}
\end{figure*}

The analysis begins with the N-CMAPSS DS03 test set, comprising six engines, $438$ flight cycles, and $425{,}156$ samples. Figure~\ref{fig:ncm_inference} summarizes how the trained autoencoder is used to reconstruct the twelve sensor variables and extract the latent health representation. Reconstruction performance is assessed using $R^2$ and Mean Absolute Error (MAE). The latent coordinates are then corrected for operating-condition effects and combined into a one-dimensional HI, whose agreement with the ground truth is evaluated using Pearson and Spearman correlations, MAE, and Root Mean Square Error (RMSE). These results are followed by comparisons with alternative representations and validation of the inferred physical trends. Section~\ref{subsec:latent_identifiability} examines latent identifiability and the contribution of an additional pressure measurement. The analysis then extends to a single model trained on pooled DS03--DS06 data to assess the diagnostic information retained across four degradation scenarios in Section~\ref{subsec:interpretable_diagnosis}. Finally, Section~\ref{subsec:component_map_results} examines the inferred turbine evolution in component-map coordinates.

\subsection{Reconstruction Performance of the Physics-Constrained Autoencoder} 
\label{subsec:reconstruction_results} 
\begin{figure}[] 
\centering 
\includegraphics[width=0.8\textwidth]{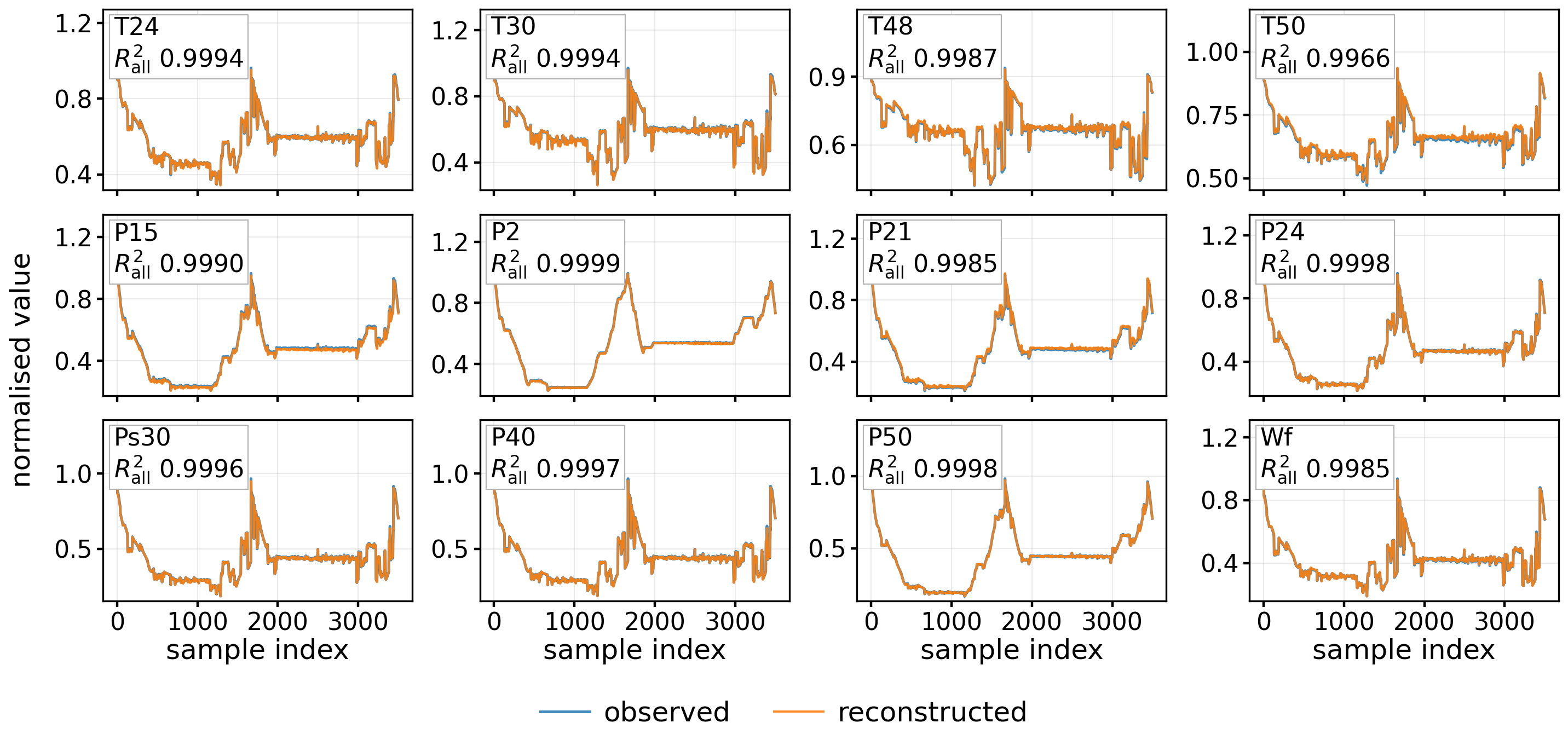} 
\caption{Measured and reconstructed trajectories for the $12$ sensor variables obtained with the proposed physics-constrained autoencoder. The reported $R^2$ values are computed over all test samples.} 
\label{fig:physics_reconstruction} 
\end{figure}

The proposed physics-constrained autoencoder achieves an overall reconstruction coefficient of determination of $R^2=0.9993$ and a MAE of $0.0036$ on the test set (engine 10, 11, 12, 13, 14, 15). Figure~\ref{fig:physics_reconstruction} compares the measured and reconstructed trajectories for the $12$ sensor variables over the test interval. The reconstructed signals follow the measured response across the different operating regimes and reproduce both the main trends and the faster variations observed within the flight profiles. The corresponding per-variable values of $R^2$ and MAE are reported in Table~\ref{tab:reconstruction_per_variable}.
The obtained reconstruction accuracy indicates that the thermodynamic computation graph can reproduce the observed engine response with only a limited error. Some discrepancies remain concentrated in specific sensor variables (i.e., $P_{21},T_{50}$), suggesting that the corresponding thermodynamic relations or component representations may require further refinement or the lack of enough observability. Nevertheless no variable exceeds a mean MAE of 0.0075, showing the consistent reconstruction ability of the architecture in all channels.

\begin{table}[b]
\centering
\caption{Per-variable reconstruction performance of the proposed physics-constrained autoencoder on the test set. Results are reported as mean $\pm$ half-range.} 
\label{tab:reconstruction_per_variable}
\renewcommand{\arraystretch}{1.04}
\setlength{\tabcolsep}{4.5pt}
\footnotesize
\begin{tabular}{lcc@{\hspace{8pt}}lcc}
\toprule
\textbf{Variable} & $\boldsymbol{R^2}$ & \textbf{MAE} &
\textbf{Variable} & $\boldsymbol{R^2}$ & \textbf{MAE} \\
\midrule

$T_{24}$ & $0.99920 \pm 0.00018$ & $0.00310 \pm 0.00044$ &
$P_{21}$ & $0.99842 \pm 0.00008$ & $0.00750 \pm 0.00015$ \\

$T_{30}$ & $0.99897 \pm 0.00039$ & $0.00371 \pm 0.00069$ &
$P_{24}$ & $0.99976 \pm 0.00006$ & $0.00208 \pm 0.00026$ \\

$T_{48}$ & $0.99884 \pm 0.00017$ & $0.00306 \pm 0.00022$ &
$P_{s30}$ & $0.99938 \pm 0.00022$ & $0.00311 \pm 0.00065$ \\

$T_{50}$ & $0.99785 \pm 0.00095$ & $0.00323 \pm 0.00071$ &
$P_{40}$ & $0.99961 \pm 0.00012$ & $0.00241 \pm 0.00031$ \\

$P_{15}$ & $0.99900 \pm 0.00005$ & $0.00598 \pm 0.00015$ &
$P_{50}$ & $0.99987 \pm 0.00007$ & $0.00189 \pm 0.00047$ \\

$P_{2}$ & $0.99986 \pm 0.00000$ & $0.00277 \pm 0.00000$ &
$W_f$ & $0.99838 \pm 0.00037$ & $0.00395 \pm 0.00065$ \\

\midrule
\multicolumn{4}{l}{\textbf{Overall}} &
$\mathbf{0.99927 \pm 0.00006}$ &
$\mathbf{0.00357 \pm 0.00015}$ \\
\bottomrule
\end{tabular}
\end{table}

The reconstruction results provide the basis for examining the inferred physical variables. An accurate reconstruction alone, however, does not demonstrate that the latent trajectories contain a suitable degradation indicator. In fact, as shown by Wen et al. \cite{wen2024residual}, in an asset such as a turbofan engine, the sensor variability is largely driven by the flight conditions, and the same contribution remains visible in the latent space before operating-condition correction.

\subsection{Removal of the Operating-Condition Contribution and 1-D HI.} \label{subsec:w_removal_results} 

Variable flight conditions strongly affect aero-engine measurements and the latent quantities inferred from them. Operating-condition mappings and residual representations have therefore been used to separate normal operating variability from degradation information~\cite{wen2024residual}. Following the procedure introduced in Section~\ref{subsec:operating_condition_correction}, we estimate this contribution using a third-order polynomial ridge model fitted to the healthy portion of each engine trajectory. The reference interval is defined as the first $15\%$ of the complete trajectory. Degradation-onset flags are used only retrospectively to verify that these intervals contain no degraded samples. The healthy interval also supplies the standard deviation used to scale each latent residual. For the HI benchmarks, coordinate orientation, selection, and fusion weights are determined on development trajectories and applied unchanged to the test engines.

\begin{figure*}[] 
\centering 
\includegraphics[width=0.8\textwidth]{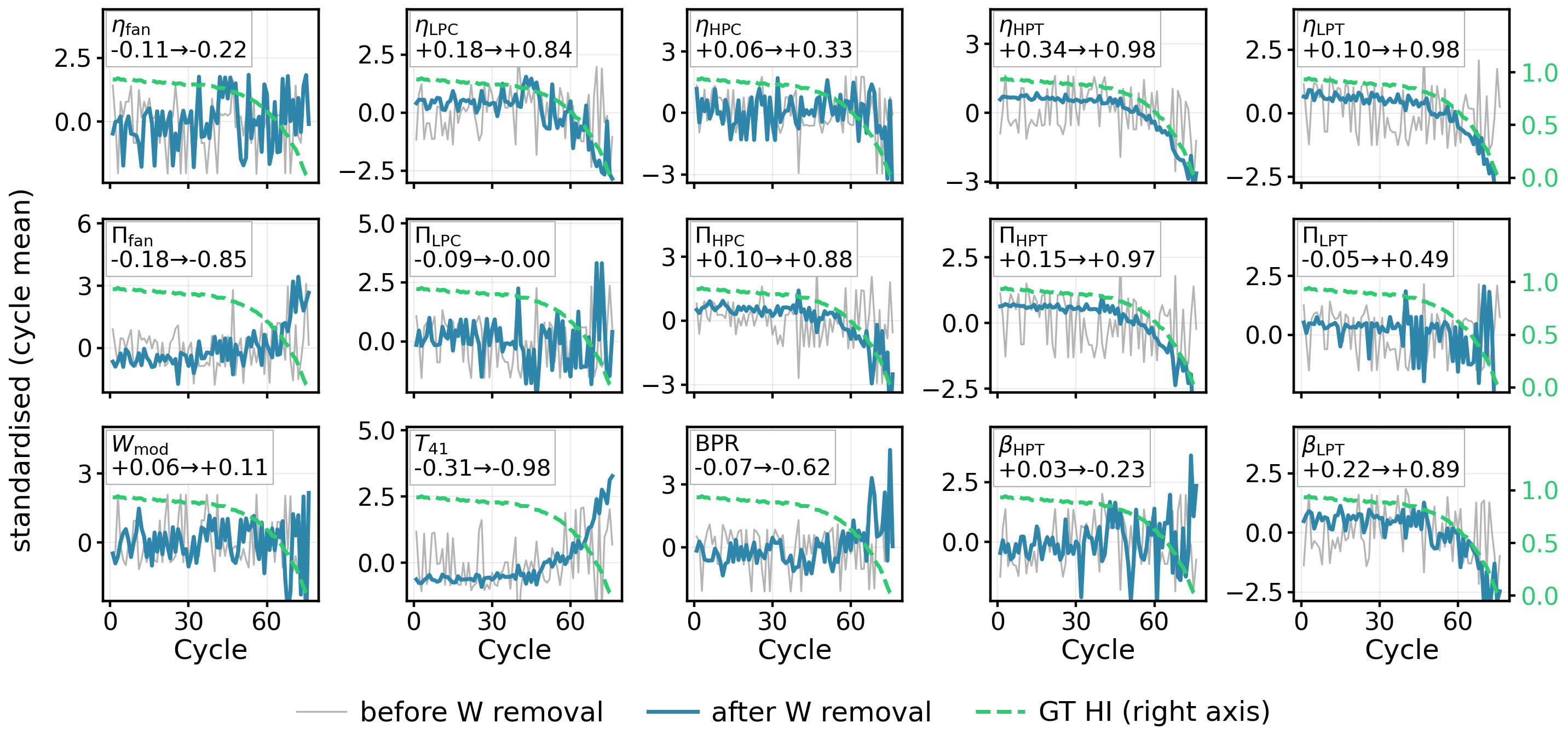} 
\caption{Latent trajectories before and after operating-condition removal for Unit~$14$. Grey lines indicate the original latent variables, blue lines their corrected residuals, and the green dashed line the engine ground-truth HI. The panel titles report the correlation before and after the correction.} 
\label{fig:latent_w_removal} 
\end{figure*}

The resulting correction substantially improves health tracking. Without correction, the HI retains a strong flight-condition contribution, with a Spearman correlation of $0.6684\pm0.1345$ against the ground truth and an RMSE of $0.1833\pm0.0845$. After correction, the correlation increases to $0.9791\pm0.0059$ and the RMSE decreases to $0.0286\pm0.0033$, approximately a sixfold reduction.
Figure~\ref{fig:latent_w_removal} illustrates the corresponding changes in the latent trajectories for a representative test engine, unit 14. After correction, the coordinates associated with turbine-inlet temperature, HPT efficiency, and LPT efficiency reach absolute correlations of $0.93$--$0.96$ with the engine ground-truth HI, compared with a maximum of $0.35$ before correction. The emergence of the two turbine-efficiency coordinates among the strongest degradation-related signals is consistent with the efficiency deterioration imposed in DS03.

To further test the proposed methodology, table~\ref{tab:w_removal_benchmark} compares the correction with alternative regressors over five runs per method. Third-order polynomial ridge achieves the highest mean HI composite score, $0.8555$, and Spearman correlation among the evaluated methods, while second-order polynomial ridge yields the lowest HI RMSE. It also explains slightly more of the healthy operating-condition dependence than the second-order polynomial model, with a W-fit $R^2$ of $0.9991$ compared with $0.9980$. Regressor selection additionally considered the need to evaluate the healthy baseline under flight conditions beyond those represented in the reference interval. Random Forest and k-NN were excluded because their predictions outside the training domain remain tied to training responses. The polynomial formulation provides an explicit mapping beyond that domain, although its extrapolation accuracy depends on the operating-condition coverage of the healthy reference data.

\begin{figure*}
    \centering
    \includegraphics[width=0.8\textwidth]{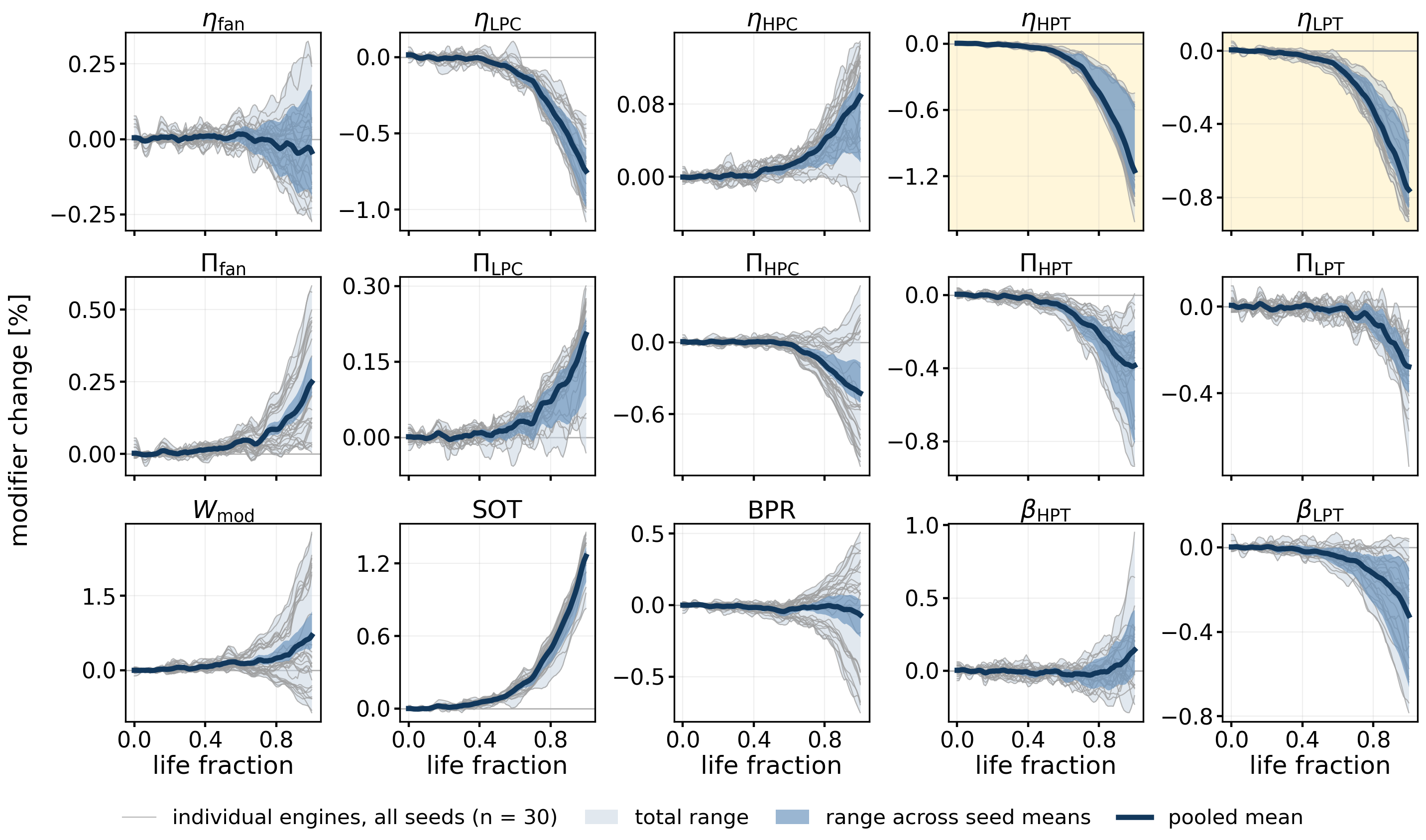}
    \caption{Cross-engine evolution of the inferred physical modifiers against normalised life fraction, over the six DS03 test engines of five independently seeded trainings (30 trajectories). Grey curves are the individual traces, the pale band their total range, the darker band the range across the five per-seed means, and the heavy curve the pooled mean. Shading marks the two coordinates degraded in DS03.}
    \label{fig:latent_consensus}
\end{figure*}

Finally, the consistency of the inferred physical trends across the six test engines is
examined in Figure~\ref{fig:latent_consensus}. To account for the different run
lengths of the engines, the trajectories are expressed against normalized life
fraction. For each latent coordinate $k$, cross-engine agreement is quantified through the mean leave-one-out Pearson correlation. The leave-one-out formulation prevents each engine from contributing to the reference trajectory against which it is
compared.

The strongest agreement is obtained for SOT,
$\eta_{\mathrm{HPT}}$, $\eta_{\mathrm{LPC}}$, and $\eta_{\mathrm{LPT}}$.
The component quantities supporting the main interpretation therefore exhibit highly repeatable trends across the test fleet, while some secondary latent coordinates remain less coherent (e.g., $BPR$, $\beta_{HPT}$).

\begin{table*}[t]
\centering

\caption{Comparison of the methods used to remove the operating-condition contribution from the latent representation. All operating-condition models are fitted using healthy samples only. For each method, four coordinates are selected and weighted using development trajectories, and the resulting HI is evaluated on the test engines. Fit times are retained from the original measurements of the unchanged fitting procedure. Each method was evaluated over five independent runs using different random seeds; results are reported as mean $\pm$ half-range over five seeds. The best and second-best results in each column are shown in bold and underlined, respectively.}

\label{tab:w_removal_benchmark}

\renewcommand{\arraystretch}{1.08}
\setlength{\tabcolsep}{4pt}
\footnotesize

\begin{tabular}{lcccccc}
\toprule
\textbf{Method} &
\textbf{Extrapolation} &
\textbf{$W$-fit $R^2$} &
\textbf{Spearman} &
\textbf{HI RMSE} &
\textbf{Composite} &
\textbf{Fit time} \\
\midrule

No removal
& --
& --
& $0.6684 \pm 0.1345$
& $0.1833 \pm 0.0845$
& $0.5601 \pm 0.1220$
& -- \\

Linear ridge
& Yes
& $0.9551 \pm 0.0335$
& $0.9409 \pm 0.0219$
& $0.0310 \pm 0.0054$
& $0.8001 \pm 0.0141$
& $\mathbf{0.8\ \mathrm{s}}$ \\

Degree-$2$ polynomial ridge
& Yes
& $0.9980 \pm 0.0009$
& $\underline{0.9775 \pm 0.0029}$
& $\mathbf{0.0265 \pm 0.0036}$
& $\underline{0.8436 \pm 0.0106}$
& $\underline{0.9\ \mathrm{s}}$ \\

\textbf{Degree-$3$ polynomial ridge}
& Yes
& $\mathbf{0.9991 \pm 0.0003}$
& $\mathbf{0.9791 \pm 0.0059}$
& $0.0286 \pm 0.0033$
& $\mathbf{0.8555 \pm 0.0133}$
& $1.1 \pm 0.1\ \mathrm{s}$ \\

MLP regressor
& Yes
& $\underline{0.9983 \pm 0.0006}$
& $0.9609 \pm 0.0035$
& $\underline{0.0271 \pm 0.0041}$
& $0.8358 \pm 0.0191$
& $85.2 \pm 9.7\ \mathrm{s}$ \\

\bottomrule
\end{tabular}
\end{table*}

After removing the operating-condition contribution, the $15$ latent residuals are aggregated at cycle level for each development and test unit and smoothed along cycles with a Savitzky–Golay filter. Finally they are evaluated individually in terms of monotonicity, trendability, prognosability and mutual information with RUL. In fact, given the unsupervised nature of the study, literature indicates this as the most reliable way of assessing unsupervised HI learning \cite{Bajarunas2024HIKnowledge}. For coordinate selection, these criteria are computed on development trajectories without using the ground-truth HI and are combined into the composite score:
$S_j
=
\frac{1}{4}
\left(
M_j+T_j+I_j+P_j
\right),
$
where $M_j$, $T_j$, $I_j$ and $P_j$ denote, respectively, the monotonicity, trendability, normalized mutual-information, and prognosability scores of the $j$-th residual latent variable. The dimensions are ranked according to $S_j$, and the first $k$ variables are retained. The final HI is obtained as their criteria-weighted combination,

\begin{equation}
\mathrm{HI}(c)
=
\frac{
\displaystyle
\sum_{j\in\mathcal{K}}
S_j\,\widetilde{z}_j(c)
}{
\displaystyle
\sum_{j\in\mathcal{K}} S_j
},
\qquad
\mathcal{K}
=
\left\{
j_1,\ldots,j_k
\right\},
\label{eq:hi_weighted_fusion}
\end{equation}

where $\widetilde{z}_j(c)$ denotes the corrected, healthy-standard-deviation-scaled, smoothed and cycle-aggregated value of the $j$-th latent variable at cycle $c$, oriented using the sign determined on development trajectories.  To determine how many coordinates to retain, the top-$k$ coordinates are fused for $k=1,\ldots,10$ and the resulting HI is evaluated on the development set. Averaged over five training seeds, the composite score is $0.8452$ with four coordinates and $0.8516$ with five. We retain four coordinates for a more compact representation, accepting the small reduction in score. The coordinate ranking, orientation, and fusion weights determined on development data are then kept fixed when constructing the HIs of the test engines.

Notably, consistent with the degradation scenario, three quantities are retained in every seed: the turbine-inlet temperature level $T_{41}$, the LPT efficiency $\eta_{\mathrm{LPT}}$, and the HPT efficiency $\eta_{\mathrm{HPT}}$; the fourth coordinate varies across seeds.
This selection is based on the evolution of the latent coordinates with cycle index and remaining useful life, without using information about the degraded components. The emergence of the two turbine efficiencies as the highest-ranked component-efficiency quantities is therefore consistent with the HPT- and LPT-efficiency degradation imposed in DS03. Their selection supports the ability of the physical latent representation to associate degradation-sensitive trends with the affected engine section.
The high ranking of $T_{41}$ captures a complementary aspect of engine degradation. Its increase can be traced back to the additional fuel and higher turbine-inlet temperature required to maintain thrust as component performance deteriorates. This coordinate therefore describes the engine's operating response to degradation, while the turbine-efficiency coordinates are directly associated with the imposed deterioration mechanisms.

The resulting HI reaches a monotonicity of $0.627$, a trendability of $0.988$, a normalized mutual-information score of $0.807$, and a composite score of $0.856$.
When compared to the ground truth, its agreement with the ground-truth HI is characterized by a Pearson correlation of $0.996$, a Spearman correlation of $0.979$, an RMSE of $0.029$, and an MAE of $0.023$. Table~\ref{tab:physics_hi_assembly} summarizes the corresponding metrics.

\begin{table}[]
\centering
\caption{Health-index construction and test performance for the proposed physics-constrained representation, with coordinate orientation, selection, and fusion weights determined on development trajectories. Results are reported as mean $\pm$ half-range over five seeds.}
\label{tab:physics_hi_assembly}
\renewcommand{\arraystretch}{1.05}
\setlength{\tabcolsep}{4pt}
\footnotesize
\begin{tabular}{lc@{\hspace{10pt}}lc}
\toprule
\textbf{Quantity} & \textbf{Value} &
\textbf{Quantity} & \textbf{Value} \\
\midrule

Monotonicity
& $0.6272 \pm 0.0503$
& Trendability
& $0.9879 \pm 0.0062$ \\

Prognosability
& $0.9998 \pm 0.0005$
& Mutual information
& $0.8072 \pm 0.0134$ \\

Composite score
& $0.8555 \pm 0.0133$
& Spearman with GT
& $0.9791 \pm 0.0059$ \\

Pearson with GT
& $0.9958 \pm 0.0016$
& MAE
& $0.0233 \pm 0.0028$ \\

RMSE
& $0.0286 \pm 0.0033$
& & \\

\bottomrule
\end{tabular}
\end{table}

\subsection{Benchmarking with Alternative Representations} 

\begin{figure*}
    \centering
    \includegraphics[width=\textwidth]{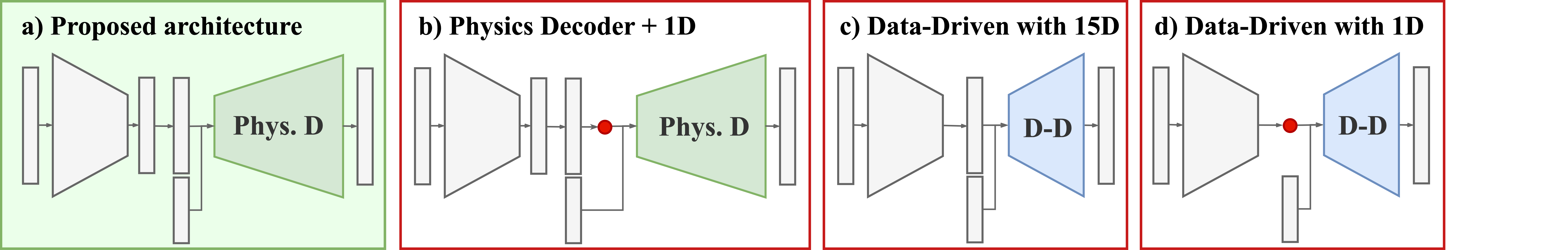}
    \caption{Overview of the benchmarking framework used to compare the proposed physics-constrained representation with the considered data-driven and reduced-order baselines.}
    \label{fig:benchmark_framework}
\end{figure*}

\label{subsec:representation_benchmark}
The proposed representation is compared with other possible HI learning techniques starting from simple linear strategies all the way to full data-driven autoencoders. In particular the different strategies, reported in Figure \ref{fig:benchmark_framework}, are: PCA representation with 1 dimension, a fully data-driven autoencoder with 15 latent dimensions, a fully data-driven autoencoder with 1 latent dimensions, a reduced-order variant of the physics-constrained autoencoder with one neuron in the latent space, and a supervised feed-forward regressor. It has to be noted that the downstream processing is identical for every unsupervised representation. Further details on the benchmarking details are provided in Appendix \ref{app:benchmark_baselines}. The declared objective is not to show that the proposed method is performing better than other baselines but that the proposed solution is able to perform reconstruction and HI learning in a competitive way with the added bonus of interpretability.

\begin{figure*}
    \centering
    \begin{subfigure}[t]{0.4\textwidth}
        \centering
        \includegraphics[width=\textwidth]{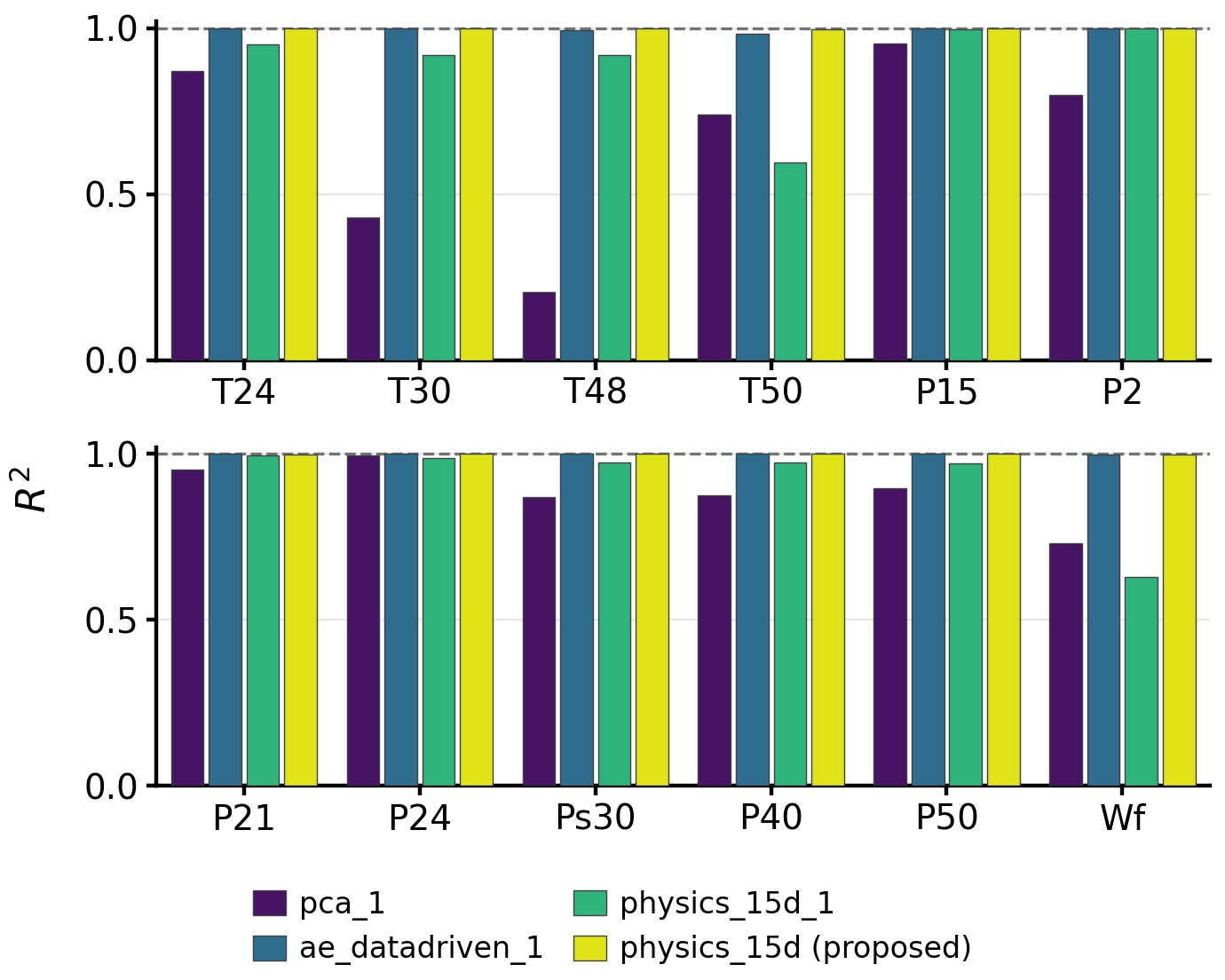}
        \caption{$R^2$}
        \label{fig:reconstruction_comparison_r2}
    \end{subfigure}
    \hfill
    \begin{subfigure}[t]{0.4\textwidth}
        \centering
        \includegraphics[width=\textwidth]{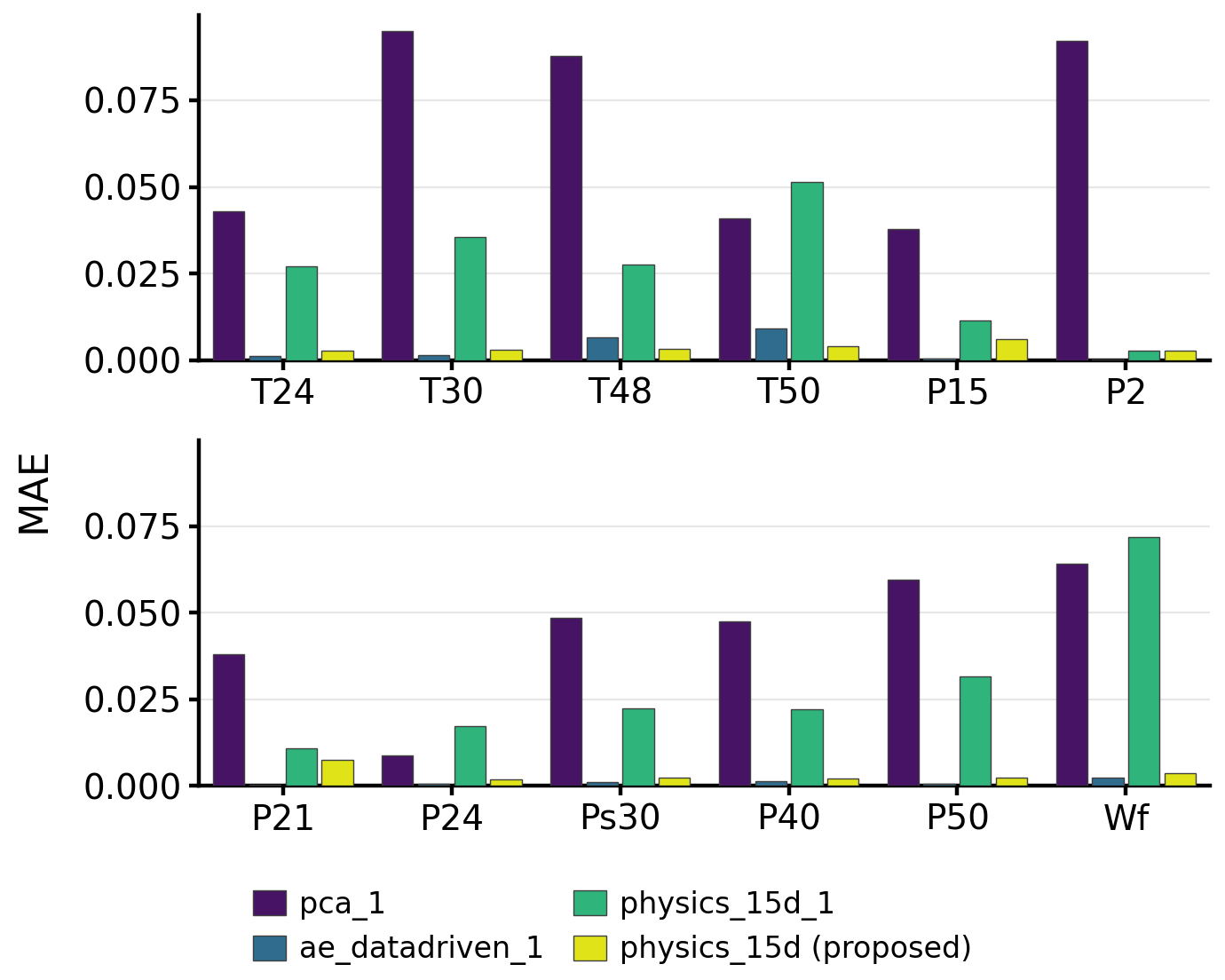}
        \caption{MAE}
        \label{fig:reconstruction_comparison_mae}
    \end{subfigure}
    \caption{Channel-wise reconstruction performance obtained by the considered representations over all test samples, reported in terms of (a) coefficient of determination $R^2$ and (b) mean absolute error (MAE).}
    \label{fig:reconstruction_comparison}
\end{figure*}

\paragraph{Reconstruction performance.}
Figure~\ref{fig:reconstruction_comparison} reports the channel-wise reconstruction performance, while Table~\ref{tab:representation_benchmark} summarizes the results over the complete test set. The 15-dimensional data-driven autoencoder obtains $R^2 = 0.9993$ and an MAE of
$0.0021$, and an identical performance is obtained with a single latent neuron,
showing how much of the reconstruction the unconstrained architecture recovers
from one coordinate. The proposed physics-constrained autoencoder reaches
$R^2 = 0.9993$ with an MAE of $0.0036$: constraining every latent coordinate to
act through fixed thermodynamic relations therefore costs no measurable
reconstruction accuracy at the level of $R^2$, and the residual difference in MAE
is the small price of the equation-based decoder. The effect of further constraining the
representation is instead more pronounced: reducing the encoder bottleneck to a
single dimension decreases the overall reconstruction $R^2$ to 0.9481 and
increases the MAE to 0.0275, which indicates that the information required by the
physical decoder cannot be compressed into a single latent coordinate and supports
the use of a multidimensional physical representation.

\begin{table*}[t]
\centering
\caption{Comparison of the considered representations on the N-CMAPSS DS03
test set. Reconstruction metrics are computed over all test samples, while HI
metrics are computed over the $438$ flight cycles. The column $k$ indicates the
number of latent coordinates retained for HI construction. For one-dimensional
representations, $k=1$ by construction; $k$ is not applicable to the supervised
baseline. For all unsupervised representations, coordinate orientation and, where applicable, coordinate selection and fusion weights are determined on development trajectories and applied unchanged to the test engines. Results with spread are reported as mean $\pm$ half-range over five seeds. The best and second-best results in each column are shown in bold and
underlined, respectively.}
\label{tab:representation_benchmark}
\renewcommand{\arraystretch}{1.08}
\setlength{\tabcolsep}{3.5pt}
\footnotesize

\begin{tabular}{lcccccc}
\toprule
\textbf{Representation} &
$\boldsymbol{k}$ &
\textbf{Recon. $R^2$} &
\textbf{Recon. MAE} &
\textbf{Spearman} &
\textbf{HI RMSE} &
\textbf{Composite} \\
\midrule

PCA, $1$ component
& 1
& 0.83108
& 0.05527
& \underline{0.9643}
& $\mathbf{0.0286}$
& 0.8266 \\

Data-driven AE, $1$-D
& 1
& $\mathbf{0.99928 \pm 0.00000}$
& $\mathbf{0.00208 \pm 0.00001}$
& $0.7545 \pm 0.3664$
& $0.1300 \pm 0.1691$
& $0.6758 \pm 0.2291$ \\

Data-driven AE, $15$-D
& 4
& $\mathbf{0.99928 \pm 0.00000}$
& $\mathbf{0.00208 \pm 0.00001}$
& $0.9525 \pm 0.0389$
& $\underline{0.0305 \pm 0.0172}$
& $\underline{0.8392 \pm 0.0599}$ \\

Physics AE, bottleneck $1$
& 1
& $0.94811 \pm 0.00074$
& $0.02745 \pm 0.00028$
& $0.9428 \pm 0.0170$
& $0.0398 \pm 0.0077$
& $0.8148 \pm 0.0147$ \\

\textbf{Physics AE, $15$-D (proposed)}
& 4
& $\underline{0.99927 \pm 0.00006}$
& $\underline{0.00357 \pm 0.00015}$
& $\mathbf{0.9791 \pm 0.0059}$
& $\mathbf{0.0286 \pm 0.0033}$
& $\mathbf{0.8555 \pm 0.0133}$ \\

Supervised baseline$^\dagger$
& --
& --
& --
& 0.8530
& 0.0820
& 0.6114 \\

\bottomrule
\end{tabular}

\vspace{2pt}
\raggedright
\scriptsize
$^\dagger$ The supervised baseline uses leave-one-unit-out training and evaluation within the designated test fleet and is not directly comparable with the development-trained representations.
PCA and the supervised baseline are deterministic; therefore, no spread is
reported.
\end{table*}

\paragraph{Health-index performance.}
A different picture emerges when the representations are evaluated according to the quality of the resulting HI. The proposed physics-constrained representation achieves the highest mean Spearman correlation, $0.9791$, and composite score, $0.8555$. Its HI RMSE is $0.0286$, compared with $0.0305$ for the 15-D data-driven AE, whose Spearman correlation and composite score are $0.9525$ and $0.8392$, respectively. PCA achieves the same HI RMSE as the proposed representation at the reported precision. It can be seen how the variation between seeds for the one dimensional data-driven architecture is very high compared to the other architectures, underscoring the instability of that solution. The other data-driven architectures still show greater variability than the proposed version. The PCA one-dimensional representation has a relatively poor reconstruction performance ($R^2=0.8311$), yet still reaches a Spearman correlation of $0.9643$. 

\begin{figure*}[] 
\centering 
\includegraphics[width=\textwidth]{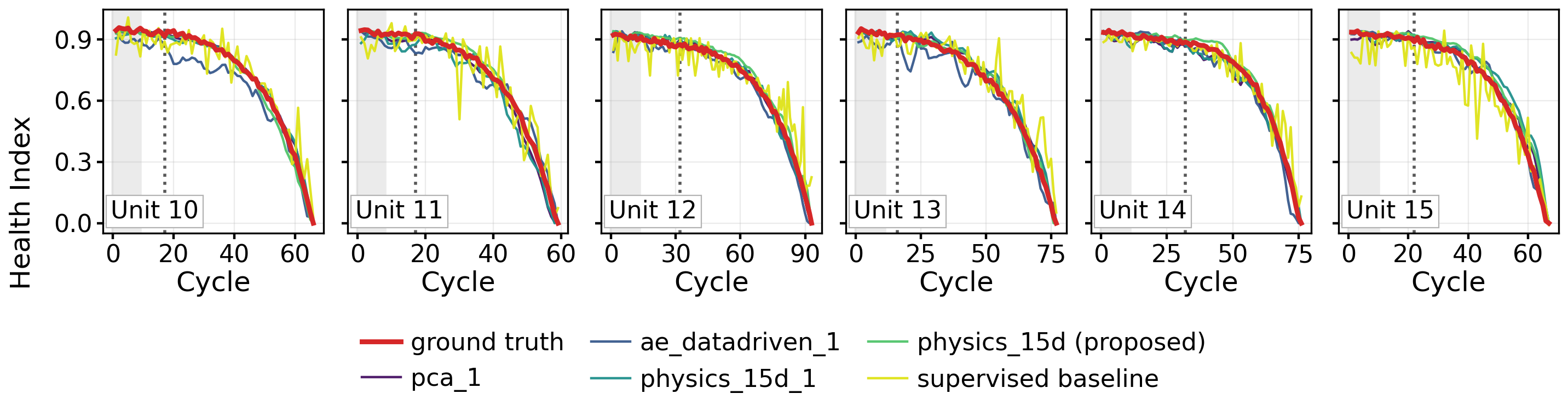} 
\caption{Estimated HIs and ground-truth HI for the six test engines. The grey regions identify the healthy reference intervals, while the dotted lines indicate the degradation-onset cycles. Almost all methodologies show a similar trend, except for the supervised model in yellow and the data-driven autoencoder with 1 dimension in the latent space. This justifies the goodness of the physics decoder while providing an explicitly physics-structured latent representation.} 
\label{fig:hi_comparison} 
\end{figure*}

The reduced-order physics-constrained architecture also confirms the importance of retaining the component-level latent structure. Compressing the physics representation to a single coordinate reduces the Spearman correlation from $0.9791$ to $0.9428$ and the composite score from $0.8555$ to $0.8148$. The performance of the proposed model therefore results from the combination of the physical decoder and the structured multidimensional latent representation, rather than from the decoder alone. Figure~\ref{fig:hi_comparison} further compares the resulting HI trajectories with the ground truth for the six test engines. The supervised baseline reaches a Spearman correlation of $0.8530$ and an RMSE of $0.0820$ and produces less regular trajectories despite being trained directly on the ground-truth HI. Since this model uses labelled degradation information, it is reported as a reference and is not directly comparable with the unsupervised representations. 

In conclusion, the benchmark therefore shows a comparable reconstruction ability with respect to fully data-driven counterparts, comparable HI accuracy compared with the strongest linear baselines, but accompanied by a substantial increase in physical interpretability.

\subsection{Latent Space Variables Validation}

The directions reported in Figure~\ref{fig:latent_consensus} are consistent with the degradation mechanisms defined for N-CMAPSS DS03, which combines HPT-efficiency degradation with LPT-efficiency and LPT-flow-capacity degradation. The corresponding efficiency coordinates exhibit clear and repeatable trends across the test engines, providing a first indication that the physical structure imposed by the decoder is reflected in the inferred latent representation. A direct assessment is possible using the model health parameters provided by N-CMAPSS in $\boldsymbol{\theta}$, which describe the degradation imposed during data generation.

\begin{figure*}
\centering
\includegraphics[width=\textwidth]{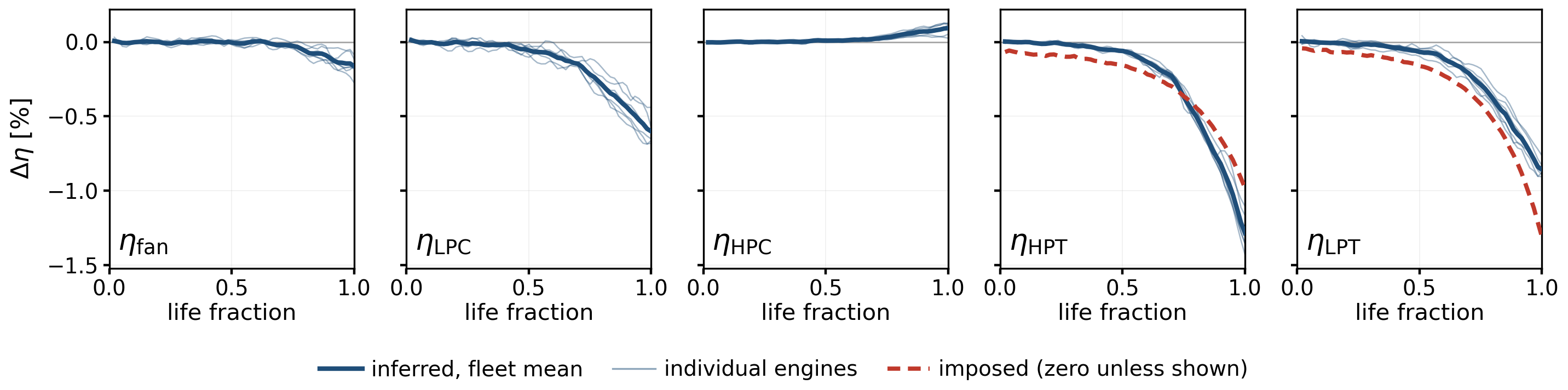}
\caption{Relative changes in the five inferred efficiency-related coordinates for DS03. Thin blue curves show individual test engines, and the thick blue curve shows their mean. Dashed red curves show the imposed HPT- and LPT-efficiency degradation. No degradation is imposed on the fan, LPC, or HPC.}
\label{fig:modifier_validation}
\end{figure*}

Figure~\ref{fig:modifier_validation} shows that the inferred HPT- and LPT-efficiency trajectories follow the same downward evolution as the imposed degradation, reproducing both its gradual initial progression and its sharper decline towards end of life. This agreement is obtained without using the reference health parameters to train the autoencoder. Across all test engines and five training seeds, the mean Spearman correlations are $0.97$ for $\eta_{\mathrm{HPT}}$ and $0.95$ for $\eta_{\mathrm{LPT}}$, with minimum values of $0.91$ and $0.86$, respectively. Together with the repeatability shown in Figure~\ref{fig:latent_consensus}, these results demonstrate that the coordinates assigned to the two turbine efficiencies retain the temporal signature of the corresponding component degradation. Quantitative interpretation of the inferred magnitudes is examined in Section~\ref{subsec:latent_identifiability}.

The comparison across all five efficiency coordinates also shows that the largest changes occur in the HPT and LPT, where efficiency degradation is imposed. The fan, LPC, and HPC coordinates exhibit smaller changes as the decoder reconstructs the response of the engine to turbine degradation. Their evolution may combine changes in component operating states with compensation among latent quantities. To examine whether these secondary trends are physically consistent with the broader engine response, the same degradation scenario is reproduced using GasTurb and TurboGrad, two independent gas-turbine performance simulation tools, in Appendix~\ref{app:gasturb}. This comparison complements the direct validation of the turbine-efficiency trends, while the coupling among latent quantities is examined in Section~\ref{subsec:latent_identifiability}.

\subsection{Physical Interpretation and Identifiability of the Latent Representation}
\label{subsec:latent_identifiability}

The physical decoder assigns each latent coordinate a predefined role in the thermodynamic equations. The interpretation of its inferred evolution depends on the degradation mechanisms represented in the parameterization and on the information provided by the measurements. Both aspects must therefore be considered when interpreting latent changes as estimates of individual physical parameters.

\paragraph{Representation of flow-capacity degradation.}

The N-CMAPSS scenarios considered in this study include degradation in both component efficiency and flow capacity. The decoder represents the five component efficiencies through dedicated latent coordinates, while flow is parameterized through the inlet modifier $W_{\mathrm{mod}}$ and the coupled thermodynamic equations. There are no separate modifiers describing changes in the flow capacity of each component. Consequently, the LPT-flow-capacity degradation cannot be recovered through corresponding component-specific coordinates.

These degradation mechanisms nevertheless affect the measured engine response that the autoencoder is trained to reconstruct. To reproduce their effects, the encoder can adjust the available latent quantities, including efficiencies, pressure ratios, and the inlet flow modifier. An inferred change in an efficiency-related coordinate may therefore partly account for a flow-capacity change that is not explicitly parameterized in the decoder, in what it is often called "spillover" or "smearing". This distinction matters when interpreting the latent trajectories, as compensation between variables may influence the magnitude of the inferred efficiency changes.

\paragraph{Identifiability of the latent quantities.}

The available measurements do not constrain all latent quantities independently. Changes in different quantities can produce similar sensor responses, allowing one latent variable to partly compensate for another during reconstruction. To examine this ambiguity, we used the decoder Jacobian, which describes how the reconstructed measurements respond to small changes in each latent coordinate. Singular value decomposition (SVD) of the mean Jacobian was then used to distinguish independent combinations of latent changes that affect the measurements from combinations whose effects cancel to first order. This analysis follows common sensitivity-based approaches to parameter identifiability~\cite{Doherty2009Identifiability,Stigter2015Identifiability}.

For all five independently trained DS03 models, the mean Jacobian has nine nonzero singular values, giving a rank of 9 and a six-dimensional null space. The evaluated mean Jacobian therefore distinguishes nine independent combinations of the fifteen latent coordinates, leaving six combinations unconstrained in this linearized analysis. This does not mean that nine individual physical parameters can be uniquely recovered. The coordinate-level analysis in Appendix~\ref{app:observability} shows that the compressor pressure ratios and HPC efficiency have negligible contributions to the unconstrained directions, while the two cooling fractions are particularly weakly constrained. The remaining quantities exhibit varying degrees of ambiguity.
The latent trajectories therefore provide physically structured information on degradation, with component-level interpretation guided by the distinguishability of their effects on the measurements.

\paragraph{Effect of the inter-turbine pressure measurement.}

The inter-turbine pressure $P_{45}$ provides information on how the total expansion is divided between the HPT and LPT. In the standard measured sensor set, this pressure is unavailable. However, the exit pressure $P_{50}$ constrains the product $\Pi_{\mathrm{HPT}}\Pi_{\mathrm{LPT}}$, while the temperatures $T_{48}$ and $T_{50}$ help distinguish the individual turbine contributions. An additional pressure measurement between the turbines could therefore improve their separation.

To examine this possibility, the virtual $P_{45}$ variable available in N-CMAPSS was added to the DS03 experiment as a thirteenth encoder input and reconstruction target. This experiment assesses the value of an additional measurement beyond the standard sensor configuration.
First of all, adding $P_{45}$ increases the rank of the mean Jacobian from 9 to 10 across all five independently trained models. The cosine similarity between the Jacobian columns associated with the two expansion ratios decreases from $0.916$ to $0.767$, indicating that their effects on the measurements become more distinct. The corresponding similarity between the turbine efficiencies remains essentially unchanged, from $0.715$ to $0.714$. The additional pressure measurement therefore primarily improves separation of the expansion ratios.

Reconstruction performance remains comparable, with $R^2=0.99932\pm0.00010$ for the thirteen-channel model and $R^2=0.99927\pm0.00006$ for the twelve-channel model. With $k=4$ for both models and coordinate orientation, selection, and fusion weights determined on development trajectories, the HI composite score is $0.8555\pm0.0133$ with twelve channels and $0.8425\pm0.0338$ with the additional $P_{45}$ measurement, showing no improvement in the mean system-level HI score. The additional information mainly benefits the physical resolution of the representation.
The inferred distribution of degradation between the turbines also changes.
The effect on turbine-degradation attribution is quantified through the relative efficiency change,
\begin{equation}
d_X
=
100\left(
\frac{\langle\eta_X\rangle_{\mathrm{EOL}}}
{\langle\eta_X\rangle_{\mathrm{H}}}
-1
\right),
\label{eq:relative_efficiency_change}
\end{equation}
evaluated here for $X\in\{\mathrm{HPT},\mathrm{LPT}\}$.
The healthy value is averaged over the first $15\%$ of each engine trajectory, while the end-of-life value is averaged over its last three flight cycles. The resulting changes are then averaged over the six test engines and five training seeds. The ground-truth efficiency changes are obtained from \texttt{HPT\_eff\_mod} and \texttt{LPT\_eff\_mod}, contained in N-CMAPSS $\boldsymbol{\theta}$ dataset.

With twelve channels, the mean inferred changes are $-1.10\%$ for the HPT and $-0.74\%$ for the LPT. Adding $P_{45}$ changes these estimates to $-0.50\%$ and $-1.18\%$, respectively, compared with ground-truth changes of $-0.93\%$ and $-1.23\%$. The ratio of the LPT to HPT change magnitudes, calculated from the mean changes, therefore increases from $0.67$ to $2.38$, compared with the ground-truth ratio of $1.33$. Adding $P_{45}$ recovers the correct ordering of the turbine-efficiency decreases across all five seeds and brings the LPT estimate closer to the ground truth, while the HPT decrease remains underestimated.

Overall, the experiment illustrates how sensor information can improve the separation and interpretation of component-related latent changes while preserving reconstruction and system-level health tracking.

\subsection{The Basis for an Interpretable Diagnosis}
\label{subsec:interpretable_diagnosis}

The physically structured evolution of the latent variables provides the basis for assessing the degradation region, subject to the interpretation limits established in Section~\ref{subsec:latent_identifiability}. Because the coordinates of the proposed representation are associated with specific physical quantities, their relative evolution can in principle indicate not only the presence of degradation, but also the region of the engine in which it develops. In this sense, the purpose of this subsection is not to introduce a new diagnostic methodology for N-CMAPSS. Instead, it is to demonstrate that the physical structure of the learned
representation already carries diagnostic information when combined with a
deliberately simple post-processing procedure, and therefore to assess directly
the added value of the physics-constrained latent space.

In order to test this, datasets containing different degradation modes are required. For this reason, the analysis is extended beyond DS03 by three further N-CMAPSS scenarios:

\begin{itemize}
    \item \textbf{DS03}: HPT efficiency, LPT efficiency and LPT flow capacity
          (expansion system);
    \item \textbf{DS04}: fan efficiency and fan flow capacity;
    \item \textbf{DS05}: HPC efficiency and HPC flow capacity;
    \item \textbf{DS06}: LPC efficiency and flow capacity together with HPC
          efficiency and flow capacity.
\end{itemize}

These datasets place degradation in four different regions of the gas path. To really test if the model is able to distinguish between degradation modes, a single model is trained on a pooled dataset with all degradation modes. Therefore, all four datasets are pooled into a single training set and one model is
trained on it, with the same 15-dimensional latent definition introduced in
Section~\ref{sec:case_study}. The final pooled dataset is constituted by a total of 27 training and 18 test engines.
Table~\ref{tab:pooled_reconstruction} shows that the pooled model preserves the reconstruction quality of the individual ones. This is a relevant result in itself: the proposed formulation is not tuned to one degradation scenario, and a single physics-constrained model reproduces the 12 measured channels equally well no matter if it is exposed to one fault mode or to four.

\begin{table}[htbp]
\centering
\caption{Reconstruction on the test engines for single-scenario models and the
pooled model. The DS03 and pooled entries are reported as mean $\pm$ half-range
over five seeds; the remaining entries are single runs.}
\label{tab:pooled_reconstruction}
\begin{tabular}{lcccc}
\hline
Training set & Test $R^2$ & Test MAE & Jacobian rank & HP shaft residual \\
\hline
DS03 (5 seeds)
& $0.99927 \pm 0.00006$
& $0.00357 \pm 0.00015$
& 9
& $0.00111 \pm 0.00028$ \\

DS04
& $0.99930$
& $0.00339$
& 9
& $0.00116$ \\

DS05
& $0.99932$
& $0.00374$
& 9
& $0.00080$ \\

DS06
& $0.99907$
& $0.00383$
& 9
& $0.00094$ \\
\hline

DS03--DS06 pooled (5 seeds)
& $0.99912 \pm 0.00009$
& $0.00383 \pm 0.00024$
& 9
& $0.00093 \pm 0.00023$ \\
\hline
\end{tabular}
\end{table}

\begin{figure*}
  \centering
  \includegraphics[width=0.9\textwidth]{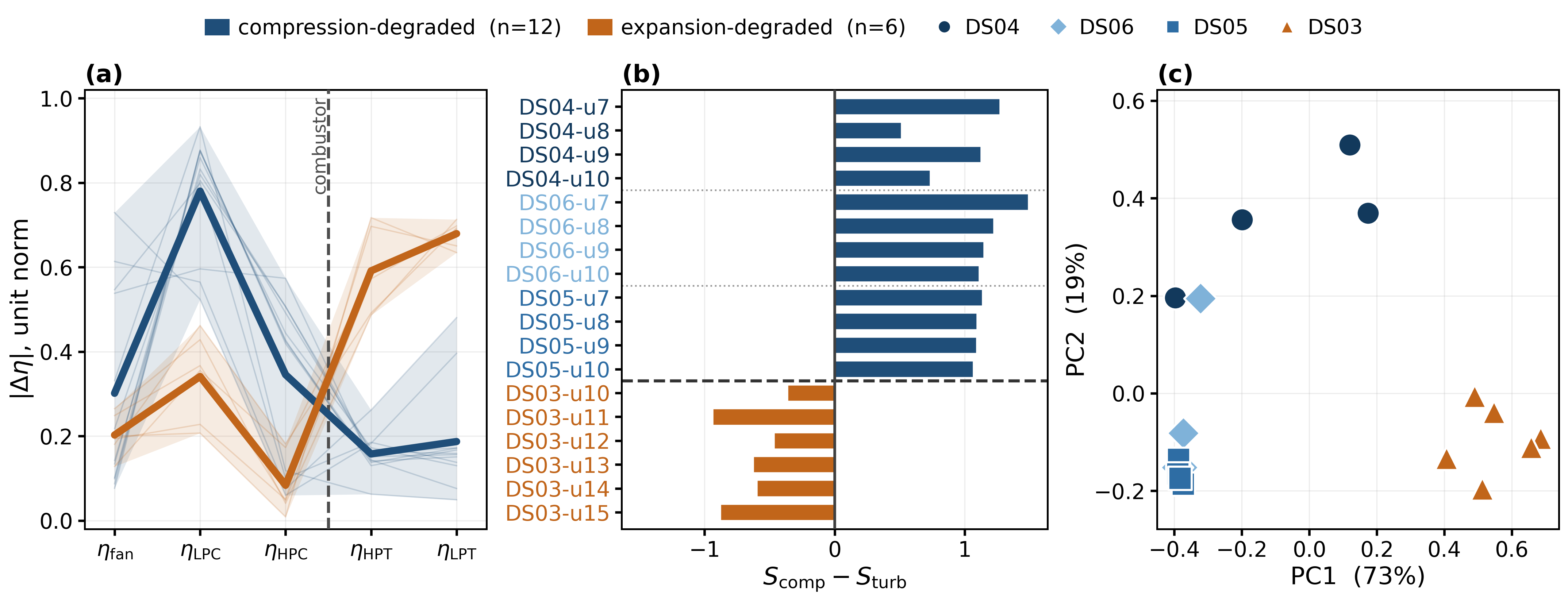}
  \caption{Degradation signatures of the 18 test engines from the pooled DS03--DS06 model. (a) Normalized changes in the five efficiency-related coordinates, with individual engine profiles and regional means with min--max bands. (b) Score difference $S_{\mathrm{comp}}-S_{\mathrm{turb}}$, correctly assigning all engines to the compression or expansion side. (c) PCA projection of the same signatures, with labels added after projection. Colour denotes the degradation region and marker shape denotes the source dataset.}
  \label{fig:diagnostic_triptych}
\end{figure*}

The diagnostic analysis uses the five efficiency-related coordinates, $\eta_{\mathrm{fan}}$, $\eta_{\mathrm{LPC}}$, $\eta_{\mathrm{HPC}}$, $\eta_{\mathrm{HPT}}$, and $\eta_{\mathrm{LPT}}$, which follow the sequence of the main components along the gas path. Their degradation-induced changes define physically structured signatures for assessing the affected engine region. As discussed in Section~\ref{subsec:latent_identifiability}, these changes may also reflect flow-capacity degradation and coupling among latent quantities. As we will demonstrate, they nevertheless already retain physically structured signatures that support discrimination between compression-side and expansion-side degradation.

For each engine, a five-dimensional degradation signature is constructed from the relative efficiency changes defined in Eq.~\eqref{eq:relative_efficiency_change}, evaluated for the fan, LPC, HPC, HPT, and LPT. The healthy and end-of-life windows are the same as those used in Section~\ref{subsec:latent_identifiability}. Before computing these changes, the latent coordinates are expressed at a common reference flight condition using Eq.~\eqref{eq:reference_condition_latent}. For the pooled model, this reference is computed as the mean over the development samples and corresponds to altitude $15\,442$ ft, Mach $0.554$, throttle-resolver angle $66.3$, and $T_2=493.7\,{}^\circ\mathrm{R}$.
The magnitudes of the five relative changes are normalized to unit Euclidean norm,
\begin{equation}
s_X
=
\frac{|d_X|}
{\sqrt{\sum_{Y\in\mathcal{C}}d_Y^2}},
\qquad
\mathcal{C}
=
\{\mathrm{fan},\mathrm{LPC},\mathrm{HPC},\mathrm{HPT},\mathrm{LPT}\}.
\label{eq:normalized_degradation_signature}
\end{equation}
The resulting signature describes the relative distribution of the inferred efficiency changes along the gas path. All three panels of Figure~\ref{fig:diagnostic_triptych}, including the diagnostic scores and PCA, use this normalized signature.

Before any test, it is useful to look at the signatures in the order in which the
components appear in the engine, which is what
Figure~\ref{fig:diagnostic_triptych}a shows. Each faint line is one test engine:
its five normalised values plotted against the physical position of the
corresponding component, from the fan at the inlet to the LPT at the exhaust,
with the combustor marked by the dashed line between $\eta_{\mathrm{HPC}}$ and
$\eta_{\mathrm{HPT}}$. The two heavy curves are the means of the twelve
compression-degraded and the six expansion-degraded engines.
The two mean profiles are very distinguishable. The
compression-degraded mean rises from $0.30$ at the fan to a maximum of $0.78$ at
$\eta_{\mathrm{LPC}}$ and then falls across the combustor to a minimum of $0.16$
at $\eta_{\mathrm{HPT}}$. The difference between the two means changes sign exactly once, between $\eta_{\mathrm{HPC}}$ and
$\eta_{\mathrm{HPT}}$: the two curves cross at the combustor. This behaviour is perfectly coherent with the expected degradation, proving the model ability to discern between degradation modes in different sections of the engine. The following paragraphs further discuss this ability.

The diagnostic assessment therefore focuses on discrimination between compression-side and expansion-side degradation. Two scores are defined by summing the magnitudes of the degradation-induced changes in the efficiency-related coordinates belonging to each side:
\begin{equation}
S_{\mathrm{comp}}
=
s_{\mathrm{fan}}+s_{\mathrm{LPC}}+s_{\mathrm{HPC}},
\qquad
S_{\mathrm{turb}}
=
s_{\mathrm{HPT}}+s_{\mathrm{LPT}}.
\label{eq:diagnostic_scores}
\end{equation}
and the degradation is assigned to the side with the larger score. The two scores group the latent efficiencies according to their position in the engine, providing a diagnostic rule that requires no additional training or fault labels.
Applied to the 18 test engines of the pooled dataset, the rule assigns every
engine to the correct side of the combustor. Figure~\ref{fig:diagnostic_triptych}b
reports the outcome engine by engine, each contributing one signed bar
$S_{\mathrm{comp}}-S_{\mathrm{turb}}$ drawn towards the side it selects and
coloured by the true region.

The grouping in Eq.~\eqref{eq:diagnostic_scores} still relies on prior knowledge
of the engine layout, since the latent coordinates are assigned in advance to
the compression and expansion sides. To check whether the same separation is
present without this information, PCA is applied directly to the
five-dimensional normalized degradation signatures of the 18 engines. This step is fully
unsupervised: the degradation-region labels are not used to compute the
projection and are added only afterwards to identify the points in
Figure~\ref{fig:diagnostic_triptych}c.

The two degradation regions separate clearly along the first principal
component, which explains 73\% of the total variance.
Compression-degraded engines occupy
$\mathrm{PC}_1 \in [-0.40,+0.17]$, while expansion-degraded engines lie in
$\mathrm{PC}_1 \in [0.41,0.69]$, with a gap of $0.232$ between the two groups.
This shows that the distinction between the two degradation regions is already present in the learned efficiency vectors, without using any information about the affected region or component. This becomes clearer when the dataset labels are added after the projection. Units from DS03, which contains turbine-related degradation, are grouped on the right side of the plot, whereas units from DS04, DS05, and DS06, associated with fan and compressor degradation, lie on the left. Within this latter group, the DS04 units, represented by dark blue circles, separate from the other compression fleets along PC2, suggesting that a more specific degradation signature may also be retained. Overall, the PCA shows that the learned efficiency coordinates retain information that separates turbine degradation from fan/compressor degradation. This analysis illustrates how the physical structure of the latent space supports diagnostic interpretation through simple post-processing; developing a dedicated diagnostic strategy lies outside the scope of this paper.

\subsection{Evolution in Component-Map Coordinates}
\label{subsec:component_map_results}

For turbofan engines, component maps are standard tools used to characterize the operating behaviour of compressors and turbines. They describe the relationship between quantities such as corrected mass flow, pressure ratio or expansion ratio, corrected speed, and efficiency, and are routinely used to assess how a component operates over its admissible operating range. Changes in component condition modify these relationships and can therefore appear as a displacement of the operating region within the map.

The proposed physics-constrained representation makes it possible to recover part of this information directly from the latent space. In particular, the inferred turbine expansion ratios and corrected flows, evaluated for each sample using its latent coordinates and recorded flight conditions, can be represented in the same coordinates used for conventional turbine maps. The resulting plots describe the operating lines occupied by each turbine over the conditions available in the data and how these lines evolve as degradation progresses. Corrected-speed lines and efficiency contours are not reconstructed. Details of the construction are provided in Appendix~\ref{app:turbine_maps}.

Figure~\ref{fig:turbine_maps_per_unit} reports this representation for the HPT and LPT of three representative test units. The grey points show the complete set of inferred operating points, while the coloured lines describe the corrected-flow--expansion-ratio relation at increasing life fractions. A systematic upward displacement is visible with increasing degradation, indicating an increase in inferred corrected flow for a given expansion ratio. For the three units shown, the increase in corrected flow amounts to $1.66\%$, $1.47\%$, and $0.99\%$ for the HPT and $1.64\%$, $1.38\%$, and $0.57\%$ for the LPT. Across all test units, the average increases are approximately $1.13\%$ and $0.64\%$, respectively. These displacements describe the operating lines recovered by the model and remain subject to the parameterization and identifiability limits established in Section~\ref{subsec:latent_identifiability}.

\begin{figure}
    \centering
    \includegraphics[width=0.7\linewidth]{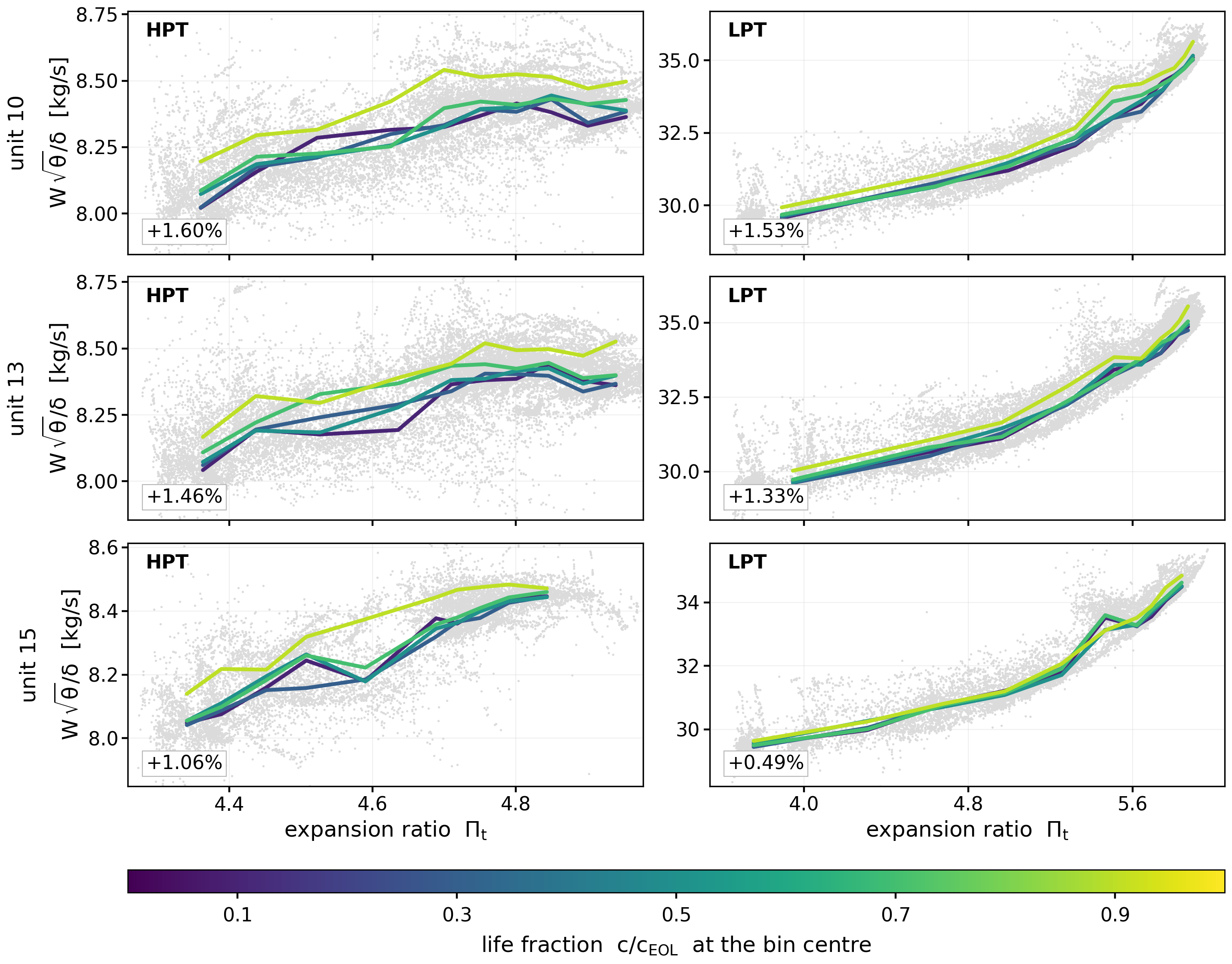}
    \caption{Degradation-induced evolution of the inferred HPT and LPT operating lines in turbine-map coordinates for three representative units with a single seed. Grey points show the complete set of inferred operating points, while coloured lines represent the corrected-flow--expansion-ratio relation at increasing life fractions. The progressive upward displacement indicates an increase in inferred corrected flow at a given expansion ratio with degradation. The percentages reported in each panel quantify the corresponding increase in corrected flow. These representations describe the evolution of operating lines in component-map coordinates and do not constitute complete turbine characteristic maps.}
    \label{fig:turbine_maps_per_unit}
\end{figure}

\section{Discussion: Physical Interpretation and Engineering Relevance}
\label{subsec:physical_interpretation}

The main contribution of the proposed architecture lies in the physical structure of its latent representation. The physics-constrained autoencoder achieves reconstruction and HI performance comparable to the considered data-driven representations while structurally associating its latent coordinates with predefined physical quantities. This structure allows the latent variables to be examined before their information is aggregated into a system-level HI. The results provide three complementary levels of interpretation: (i) component-related degradation trends in the latent space, (ii) their representation through conventional turbofan component-map coordinates, and (iii) a physically defined basis for diagnostic analysis. These results are obtained using conventional thermodynamic relations drawn from textbook turbofan-cycle analysis. Even at this level of model fidelity, the equations provide sufficient structure to recover degradation-sensitive latent trends, support area diagnostic interpretation, and express the inferred evolution in component-map coordinates.

The DS03 results illustrate how the representation captures both the affected turbine quantities and the broader thermodynamic response to degradation. The two turbine efficiencies emerge among the most informative latent quantities although the model is not provided with information on which components are degraded. The high ranking of $T_{41}$ captures a complementary aspect of engine degradation: its evolution reflects the operating response as component performance deteriorates. The degradation information is therefore expressed through coordinates associated with the affected turbine section and through quantities describing the response of the engine as a whole. The agreement of most inferred trends with the two external performance simulators provides a qualitative consistency check of this interpretation.

The diagnostic signatures and component operating lines show how this information can be examined in forms familiar to engine specialists. In the pooled experiment, simple physically grouped scores and PCA reveal a distinction between compression-side and expansion-side degradation. The component-map coordinates subsequently express the inferred evolution through relationships between corrected flow and expansion ratio. These representations complement the scalar HI by retaining information on how degradation manifests within the engine. Their engineering value lies in making the learned health evolution accessible for physical assessment. Developing a dedicated diagnostic strategy lies outside the scope of this paper.

The $P_{45}$ experiment further shows how the physical resolution of the representation can change with the sensor configuration. The additional inter-turbine pressure increases the rank of the mean Jacobian and improves separation of the turbine expansion ratios, while reconstruction and system-level HI performance remain comparable. Sensor information therefore affects which component-related changes can be distinguished, even when the existing measurements already support accurate reconstruction and health tracking. The physical decoder provides a framework in which this relationship between measurements and latent interpretation can be examined explicitly.

\section{Limitations of the Present Study}
\label{sec:limitations}

The proposed architecture embeds a set of conventional turbofan thermodynamic relations in the decoder and demonstrates that these relations can force the latent space toward physically meaningful quantities while preserving adequate reconstruction and HI performance. Nevertheless, this study still faces some limitations, explained in the next paragraphs.

\textbf{Quantitative parameter estimation.} The present study establishes the usefulness of the latent representation for health tracking and regional degradation interpretation. Quantitative estimation of individual component parameters requires further validation, with the analysis in Section~\ref{subsec:latent_identifiability} identifying where additional measurements and a richer degradation parameterization could improve their recovery.

\textbf{Model Fidelity.} A further limitation concerns the fidelity of the embedded thermodynamic model. The decoder provides a simplified representation of the engine cycle and cannot reproduce all secondary flows, component characteristics, control interactions, and nonlinear couplings represented in higher-fidelity performance models. The physical meaning recovered in the latent space is consequently bounded by the equations through which the latent variables act. Increasing the fidelity of these relations could improve the quantitative estimation and separation of individual component quantities, although this would introduce additional modelling, biases, and computational requirements. This also affect the definition of the initial set of parameters fitted by the model.

\textbf{Healthy reference and trajectory processing.} The operating-condition correction requires an initial healthy reference interval for each monitored unit. In this benchmark, its length, temporal smoothing, and per-engine HI anchoring use complete run-to-failure trajectories. Application during operation would require a healthy interval defined independently of eventual lifetime, causal processing, and an HI scale established without the monitored engine's end-of-life values.

The results therefore establish the level of interpretability that can already be obtained from a relatively compact set of conventional thermodynamic equations. The remaining limitations point primarily to the physical formulation and to the information content of the sensor set adopted in the main experiments. Future developments should investigate richer component models, additional informative measurements, and observability-aware formulations capable of improving the separation of physically related degradation effects.

\section{Conclusion}
\label{sec:conclusion}

This work introduced a physics-decoder autoencoder for interpretable health representation learning in PHM. Rather than relying on a fully trainable neural decoder, the proposed architecture embeds differentiable physical equations directly in the reconstruction path, thereby structurally associating the latent coordinates with predefined physical quantities. The resulting representation can be examined at the level of individual physical quantities before being aggregated into a system-level HI.

The methodology was demonstrated on N-CMAPSS, where the proposed model achieved a test reconstruction \(R^2\) of 0.9993 and an MAE of 0.0036 while remaining competitive with fully data-driven autoencoder representations. After removal of the operating-condition contribution, the resulting HI achieved a Spearman correlation of 0.9791 with the available ground-truth HI, with an RMSE of 0.0286 and an MAE of 0.0233. Beyond system-level health estimation, the structured latent representation retained physically organized degradation information. The inferred turbine quantities exhibited degradation trends consistent with the imposed degradation mechanisms and could be represented in conventional component-map coordinates. When a single model was trained across four degradation scenarios, the learned efficiency coordinates further enabled compression-side and expansion-side degradation to be separated without training an additional diagnostic classifier.

The results also demonstrate an important limitation of physics-structured representation learning: structural association with a physical quantity does not guarantee that the corresponding quantity is uniquely identifiable from the available measurements. In other words, a physics-constrained representation learning can make a latent health space physically structured, but the achievable physical resolution is fundamentally limited by observability. The decoder Jacobian reveals limited observability and parameter coupling for several component quantities, particularly between the fan and LPC and between the HPT and LPT. The additional \(P_{45}\) experiment further showed that introducing an informative measurement increases the observable dimension of the latent space without materially changing system-level HI performance. The absence of dedicated component-specific flow-capacity modifiers further limits quantitative attribution, since these degradation effects may be represented through other latent quantities. Physical interpretability therefore depends on the adopted parameterization and the information content of the sensor set.

Future work will investigate observability-aware latent formulations, improved thermodynamic representations, and uncertainty-aware health estimation and prognostics. Overall, the proposed framework provides a basis for moving from purely degradation-sensitive health representations toward physically structured health representations whose achievable resolution can be explicitly assessed in terms of system observability.

\section*{CRediT Authorship Contribution Statement}
\textbf{Leonardo Baldo:} Methodology, Software, Validation, Formal analysis, Investigation, Visualization, Writing - Original Draft, Writing - Review \& Editing
\textbf{Jokin Alcibar Ibarzabal:} Methodology, Software, Validation, Investigation, Visualization, Writing - Review \& Editing
\textbf{Daniel Cisneros Acevedo:} Methodology, Software, Validation, Writing - Review \& Editing
\textbf{Kai Goebel:} Conceptualization, Validation, Writing - Review \& Editing, Supervision
\textbf{Manuel Arias Chao:} Conceptualization, Methodology, Validation, Writing - Review \& Editing, Supervision, Funding acquisition

\section*{Declaration of Competing Interest}
The authors declare that they have no known competing financial interests or personal relationships that could have appeared to influence the work reported in this paper.

\section*{Code and Data Availability Statement}
Data sharing is not applicable to this article as no new data were created or generated. The N-CMAPSS dataset analysed in this study is publicly available \cite{AriasChao2021NCMAPSS}. The code needed to reproduce the analyses will be released in a public repository upon acceptance of the paper.

\section*{Declaration of generative AI and AI-assisted technologies in the manuscript preparation process}
During the preparation of this work, the authors used ChatGPT and Claude to assist with language editing, improving readability, and refining the presentation of the manuscript. After using these tools, the authors reviewed and edited the content as needed and take full responsibility for the content of the published article.


\bibliographystyle{unsrt}
\bibliography{References}

\appendix

\section{Construction of the Turbine Operating Lines in Component-Map Coordinates}
\label{app:turbine_maps}

The HPT and LPT representations reported in Section~\ref{subsec:component_map_results}
are constructed directly from quantities evaluated by the physical decoder. Corrected
mass flow is referred to ISA sea-level conditions,
$T_{\mathrm{ref}}=288.15~\mathrm{K}$ and
$P_{\mathrm{ref}}=101.325~\mathrm{kPa}$, according to

\begin{equation}
\theta_i = \frac{T_i}{T_{\mathrm{ref}}},
\qquad
\delta_i = \frac{P_i}{P_{\mathrm{ref}}},
\qquad
W_{c,i} = W_i\frac{\sqrt{\theta_i}}{\delta_i}.
\end{equation}

The operating lines are evaluated using each sample's original latent coordinates $\mathbf{z}_t$ and recorded operating conditions $\mathbf{w}_t$. The grey cloud therefore represents the sampled operating points of each engine. In contrast, the degradation signatures in Section~\ref{subsec:interpretable_diagnosis} use the transported coordinates $\mathbf{z}_{t,\mathrm{ref}}$ together with the fixed conditions $\mathbf{w}_{\mathrm{ref}}$. For the operating lines, early- and late-life comparisons are made within common expansion-ratio bins, as described below. For the two turbines, the correction is applied at the corresponding component inlet:

\begin{equation}
W_{c,\mathrm{HPT}}
=
W_{415}
\frac{\sqrt{T_{415}/T_{\mathrm{ref}}}}
     {P_{415}/P_{\mathrm{ref}}},
\qquad
W_{c,\mathrm{LPT}}
=
W_{46}
\frac{\sqrt{T_{46}/T_{\mathrm{ref}}}}
     {P_{46}/P_{\mathrm{ref}}}.
\end{equation}

The abscissa is the turbine expansion ratio, obtained directly from its latent coordinate as $\Pi_X=\Pi_X^0\exp(0.05z_{\Pi_X})$, with $X\in\{\mathrm{HPT},\mathrm{LPT}\}$. The ordinate is the corrected mass flow calculated from the corresponding decoder flow, pressure, and temperature.

For each engine, the samples are divided into life-fraction intervals $L_\ell$ and into common expansion-ratio bins. Since the corrected flow $W_c$ depends on the turbine expansion ratio $\Pi_t$, a direct comparison between early- and late-life values could be affected by differences in the operating conditions experienced in the two periods. Using the same expansion-ratio bins for all life intervals therefore allows the evolution of $W_c$ to be evaluated at comparable values of $\Pi_t$.

For each life interval $\ell$, an operating line is obtained by computing the median corrected flow in each expansion-ratio bin:

\begin{equation}
\bar W_{c,\ell,b}
=
\operatorname{median}
\left\{
W_{c,i} :
\Pi_{t,i}\in[e_{b-1},e_b),
\;
\tau_i\in L_\ell
\right\},
\qquad
\bar\Pi_{t,b}
=
\frac{e_{b-1}+e_b}{2},
\end{equation}

where $\tau$ is the normalized life fraction and $\bar\Pi_{t,b}$ is the center of bin $b$.

The displacement of the operating line over the engine life is then quantified by comparing the corrected flow in the first and last life intervals at the same expansion-ratio bins:

\begin{equation}
\Delta_c
=
\frac{100}{|B|}
\sum_{b\in B}
\left(
\frac{\bar W_{c,n_\ell,b}}
     {\bar W_{c,1,b}}
-1
\right),
\end{equation}

where $B$ contains only the bins for which corrected-flow values are available in both the first and last life intervals. Thus, $\Delta_c$ represents the mean percentage change in corrected flow from early to late life at comparable turbine expansion ratios.

\section{Benchmarking Baselines}
\label{app:benchmark_baselines}

\paragraph{Data-driven autoencoder.}
The data-driven autoencoder uses the same encoder structure as the proposed model.
The 12 normalized sensor measurements are processed by a first
$\mathrm{Dense}(256,\mathrm{ReLU})$ layer, followed by four
$\mathrm{Dense}(256,\mathrm{ReLU})$ layers, and a linear output layer of dimension $n_z$. The operating conditions are not provided to the encoder.
Two configurations are considered, with $n_z=15$ and $n_z=1$.

The physical decoder is replaced by a fully trainable FiLM-conditioned neural decoder. Denoting the four operating conditions by $\mathbf{w}$, the modulation is defined as

\begin{equation}
\mathrm{FiLM}(\mathbf{h},\mathbf{w})
=
\boldsymbol{\gamma}(\mathbf{w})\odot\mathbf{h}
+
\boldsymbol{\beta}(\mathbf{w}),
\end{equation}

where $\boldsymbol{\gamma}(\mathbf{w})$ and
$\boldsymbol{\beta}(\mathbf{w})$ are affine mappings from
$\mathbb{R}^{4}$ to $\mathbb{R}^{256}$. The decoder is composed of an initial 256-unit layer, four FiLM-modulated 256-unit blocks, a 245-unit fully connected layer, and a linear 12-dimensional reconstruction layer. FiLM conditioning was adopted to limit the direct unbalance between operating conditions and reconstructed measurements observed when $\mathbf{w}$ is concatenated directly with the latent
vector. 

The data-driven models are trained using the MSE reconstruction loss, Adam with a learning rate of $10^{-3}$, a batch size of 512, gradient-norm clipping at 1.0, and a maximum of 120 epochs. Early stopping is applied to the validation loss with a patience of 25 epochs and the best weights are restored.

\paragraph{One-dimensional physics-constrained representation.}
The reduced-order physics baseline preserves the same 15-dimensional thermodynamic decoder used by the proposed model. The encoder produces a single latent coordinate, which is mapped through a trainable linear layer to the 15 physical modifiers before entering the physical decoder.

\section{Further Latent Space Validation}
\label{app:gasturb}
In an effort to provide additional validation independent of the N-CMAPSS simulation environment, the degradation scenario was reproduced using GasTurb and TurboGrad gas turbine performance simulations tools for design and off-design analysis~\cite{10.1115/95-GT-147,rath2023investigation, acevedo2026turbograd}. A generic two-spool turbofan configuration was used to provide an external qualitative consistency check of the degradation-induced trends. The simulators are not intended to reproduce the N-CMAPSS engine quantitatively. Instead, the comparison focuses on the direction of change of the inferred physical quantities. Table~\ref{tab:gasturb_validation} shows that the changes inferred by the proposed model follow for most of the channels the same qualitative directions as those obtained with GasTurb and TurboGrad when HPT-efficiency, LPT-efficiency, and LPT-flow-capacity degradation are imposed. Compared to GasTurb, the discrepancies are linked to $\Pi_{\mathrm{HPC}}$, $\Pi_{\mathrm{LPT}}$, $\eta_{fan}$, and the cooling flows. TurboGrad shows discrepancies in $\eta_{\mathrm{HPC}}$ and $\mathrm{BPR}$, alongside those in $\Pi_{\mathrm{HPC}}$ and $\eta_{\mathrm{fan}}$. The observed discrepancies may also partly result from differences in simulator's control strategy. GasTurb and TurboGrad both employ \(N_c\)-based control, whereas the N-CMAPSS high-bypass class engine is expected to rely primarily on low-pressure spool/fan control. These different control assumptions can alter how degradation effects propagate through pressure ratios, flows, and bypass ratio. Moreover, the cooling-flow fractions in the simulators are kept constant, preventing a direct comparison of their inferred evolution. 

This additional analysis complements the quantitative evidence already obtained from the main benchmark. The GasTurb and TurboGrad comparison therefore addresses the component-level physical interpretation specifically, providing an independent consistency check where direct experimental validation of the inferred modifiers is not available.

\begin{table}[t]
\centering
\caption{Qualitative comparison of the degradation-induced evolution of the physical quantities inferred by the proposed model and obtained with GasTurb and TurboGrad. $\uparrow$, $\downarrow$, and $\rightarrow$ indicate an increase, decrease, and no clear systematic variation, respectively.}
\label{tab:gasturb_validation}

\renewcommand{\arraystretch}{0.95}
\setlength{\tabcolsep}{4pt}
\footnotesize

\begin{tabular}{cccc@{\hspace{10pt}}cccc}
\toprule
\textbf{Variable} & \textbf{Our model} & \textbf{GasTurb} & \textbf{TurboGrad} &
\textbf{Variable} & \textbf{Our model} & \textbf{GasTurb} & \textbf{TurboGrad} \\
\midrule

$\eta_{\mathrm{fan}}$  & $\rightarrow$ & \textcolor{red}{$\downarrow$}   & \textcolor{red}{$\downarrow$} &
$\Pi_{\mathrm{HPC}}$   & $\downarrow$  & \textcolor{red}{$\uparrow$}     & \textcolor{red}{$\uparrow$} \\

$\eta_{\mathrm{LPC}}$  & $\downarrow$  & $\downarrow$   & $\downarrow$ &
$\Pi_{\mathrm{HPT}}$   & $\downarrow$  & $\downarrow$   & $\downarrow$ \\

$\eta_{\mathrm{HPC}}$  & $\uparrow$    & $\uparrow$     & \textcolor{red}{$\rightarrow$} &
$\Pi_{\mathrm{LPT}}$   & $\downarrow$  & \textcolor{red}{$\uparrow$}     & $\downarrow$\\

$\eta_{\mathrm{HPT}}$  & $\downarrow$  & $\downarrow$   & $\downarrow$ &
$W_{\mathrm{mod}}$     & $\uparrow$    & $\uparrow$     & $\uparrow$ \\

$\eta_{\mathrm{LPT}}$  & $\downarrow$  & $\downarrow$   & $\downarrow$ &
$\mathrm{BPR}$         & $\rightarrow$ &  $\rightarrow$ & \textcolor{red}{$\uparrow$}\\

$\Pi_{\mathrm{fan}}$   & $\uparrow$    & $\uparrow$     & $\uparrow$ &
$T_{41}$               & $\uparrow$    & $\uparrow$     & $\uparrow$\\

$\Pi_{\mathrm{LPC}}$   & $\uparrow$    & $\uparrow$     & $\uparrow$ &
$\beta_{\mathrm{HPT}}$ & $\rightarrow$ & --             & -- \\

 & & & &
$\beta_{\mathrm{LPT}}$ & $\downarrow$  & --             & --\\

\bottomrule
\end{tabular}
\end{table}

\section{Decoder Sensitivity and Latent Identifiability}
\label{app:observability}

This appendix describes the sensitivity and identifiability analyses used to select the latent coordinates and assess their interpretation. Sensitivity measures how strongly a coordinate affects the reconstructed measurements. Identifiability additionally concerns whether this effect can be distinguished from changes in other coordinates.

\paragraph{Sensitivity and latent-space selection.}

For a given latent vector and operating condition, the decoder Jacobian is
\begin{equation}
\mathbf{J}
=
\frac{\partial\hat{\mathbf{x}}}{\partial\mathbf{z}},
\qquad
J_{ik}
=
\frac{\partial\hat{x}_i}{\partial z_k}.
\end{equation}
Each column describes the response of the reconstructed sensor variables to a small change in one latent coordinate, with the operating conditions held fixed. Its norm measures the strength of this response in normalized sensor units per unit of latent change.

\begin{figure}
    \centering
    \begin{subfigure}[t]{0.49\textwidth}
        \centering
        \includegraphics[width=0.7\linewidth]{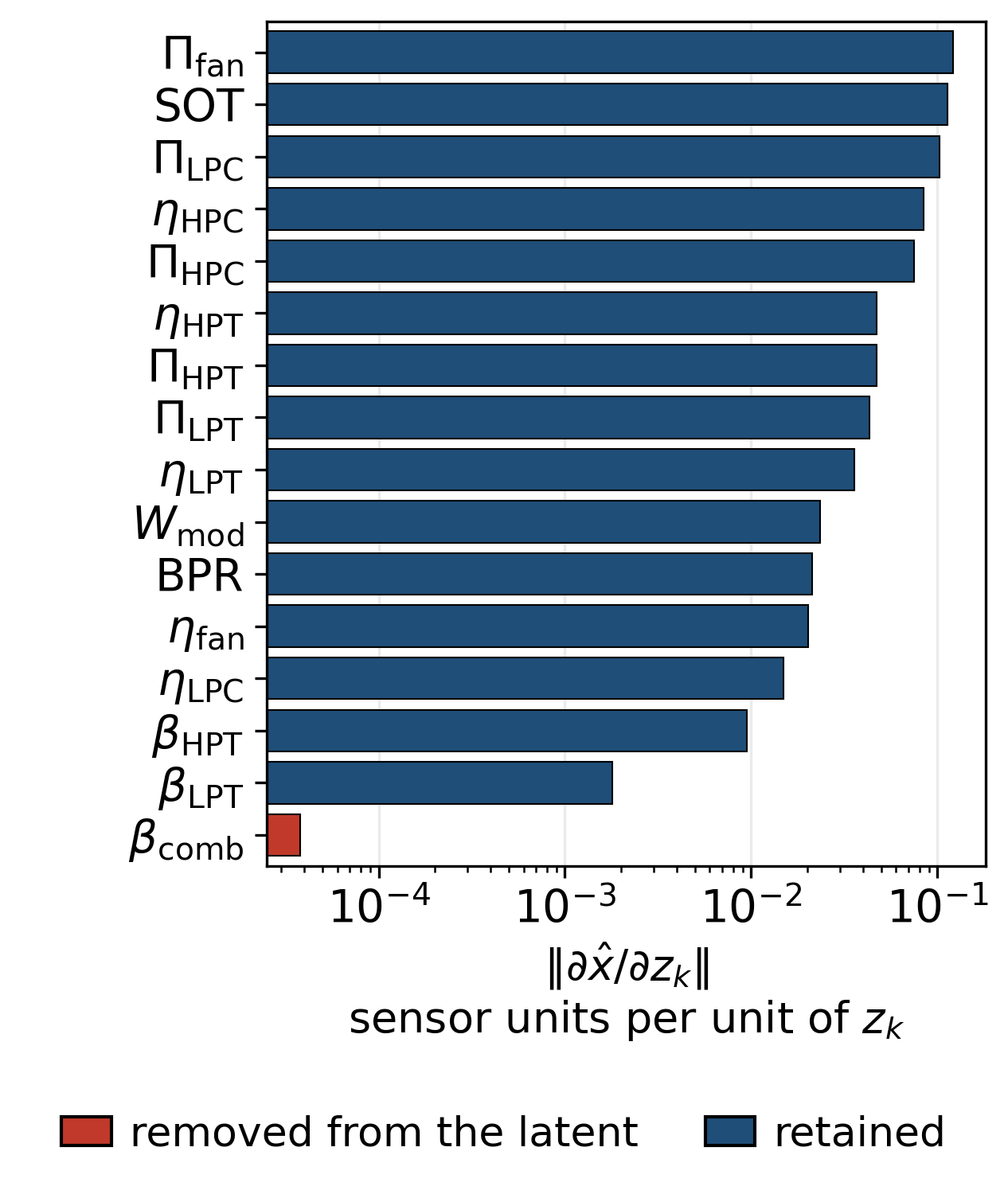}
        \caption{Latent sensitivity.}
        \label{fig:latent_observability}
    \end{subfigure}
    \hfill
    \begin{subfigure}[t]{0.39\textwidth}
        \centering
        \includegraphics[width=0.7\linewidth]{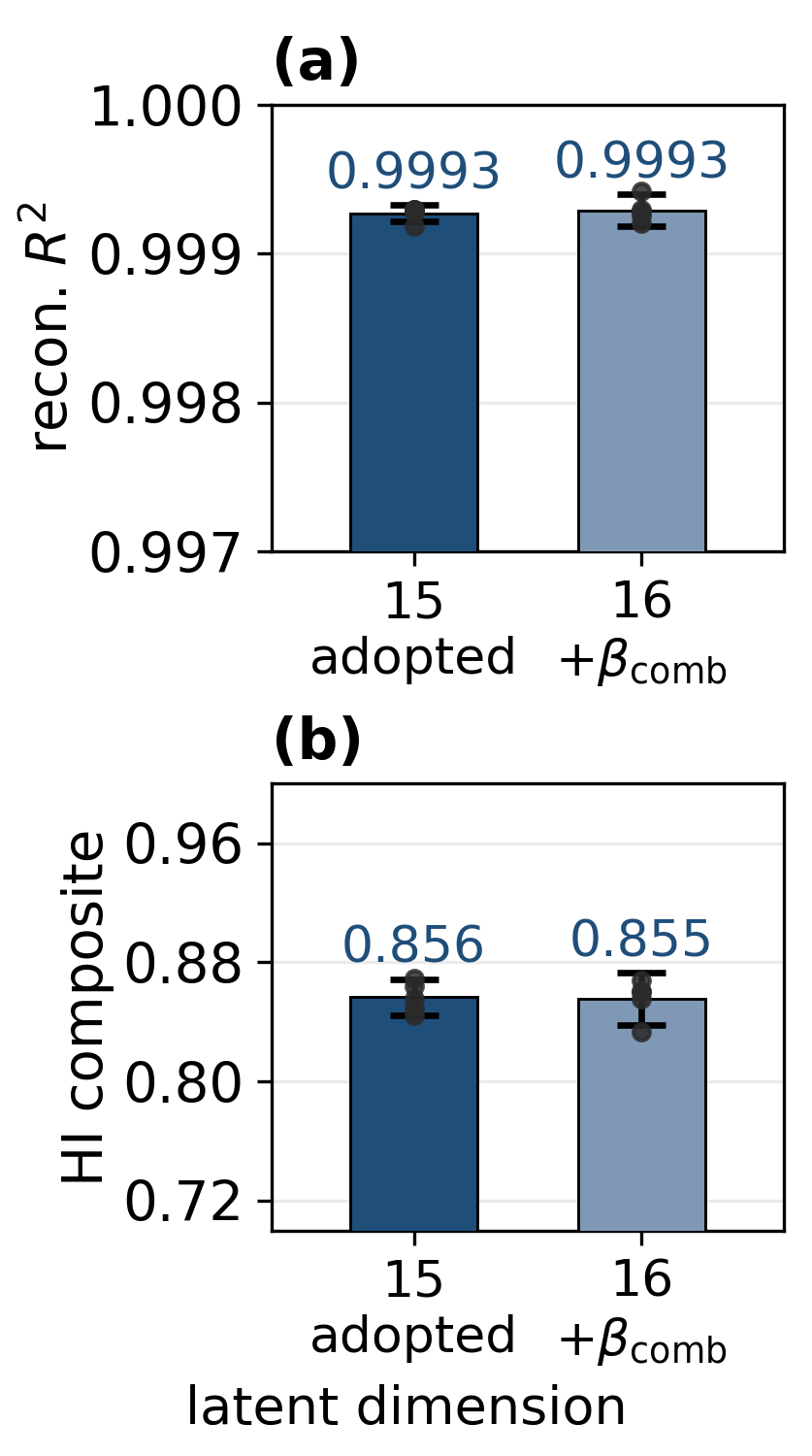}
        \caption{Cooling-coordinate ablation.}
        \label{fig:cooling_ablation}
    \end{subfigure}
    \caption{Sensitivity of the latent coordinates and effect of retaining the discarded combustor cooling fraction. (a) Norm of each coordinate's column of the decoder Jacobian, in normalized sensor units per unit of latent change, on a logarithmic axis. The retained coordinates are shown in blue and $\beta_{\mathrm{comb}}$ in red. (b) Reconstruction $R^2$ and HI composite score for the adopted 15-dimensional representation and the 16-dimensional representation retaining $\beta_{\mathrm{comb}}$. Bars show means over five independently seeded training runs, error bars show the spread across seeds, and dots show individual runs. Reintroducing the coordinate provides no improvement in either metric.}
    \label{fig:latent_selection}
\end{figure}

The initial 16-dimensional representation was examined using these column norms and the corresponding reconstruction-loss gradients, reported in Figure~\ref{fig:latent_observability} and Table~\ref{tab:observability}. The combustor cooling fraction $\beta_{\mathrm{comb}}$ has a Jacobian-column norm of $3.77\times10^{-5}$, compared with $1.80\times10^{-3}$ for the next lowest coordinate, $\beta_{\mathrm{LPT}}$. The ablation in Figure~\ref{fig:cooling_ablation} further shows that retaining $\beta_{\mathrm{comb}}$ provides no improvement in reconstruction or HI quality. This coordinate was therefore fixed at its nominal value, yielding the adopted 15-dimensional representation.

\paragraph{Construction of the mean Jacobian.}

The Jacobian is evaluated at $1\,924$ operating points sampled using a grid defined by quartile intervals of altitude, Mach number, and throttle-resolver angle, with samples drawn from each occupied cell. At each point, derivatives of the twelve normalized reconstruction targets with respect to the fifteen latent coordinates are approximated by central finite differences, one coordinate at a time, using a step of $10^{-3}$ in latent units. The decoder evaluations are performed in single precision, consistent with model training, while the Jacobian matrices are stored and averaged in double precision. The SVD is also computed in double precision. Finite differences provide a common sensitivity calculation across the decoder variants.

The matrix entering the decomposition is the arithmetic mean,
\begin{equation}
\overline{\mathbf{J}}
=
\frac{1}{N}\sum_{i=1}^{N}\mathbf{J}_i
\in\mathbb{R}^{12\times15},
\qquad N=1\,924.
\end{equation}
Numerical rank is determined by counting singular values satisfying $\sigma_j>10^{-3}\sigma_1$.

\paragraph{Singular value decomposition.}

The retained coordinates were then examined jointly through SVD of the mean decoder Jacobian,
\begin{equation}
\overline{\mathbf{J}}
=
\mathbf{U}\boldsymbol{\Sigma}\mathbf{V}^{\top},
\qquad
\overline{\mathbf{J}}\in\mathbb{R}^{12\times15}.
\end{equation}
In the full decomposition, $\mathbf{U}\in\mathbb{R}^{12\times12}$ and $\mathbf{V}\in\mathbb{R}^{15\times15}$ contain orthonormal output and latent directions, respectively, while $\boldsymbol{\Sigma}\in\mathbb{R}^{12\times15}$ contains the singular values.

Each column $\mathbf{v}_j$ of $\mathbf{V}$ describes a combination of latent-coordinate changes. For the first twelve directions,
\begin{equation}
\overline{\mathbf{J}}\mathbf{v}_j
=
\sigma_j\mathbf{u}_j,
\qquad j=1,\ldots,12,
\end{equation}
where $\mathbf{u}_j$ is the corresponding column of $\mathbf{U}$. The singular value $\sigma_j$ measures the strength of the resulting output response. A nonzero singular value therefore identifies a latent direction that affects the measurements. A zero singular value identifies a direction that produces no output change in the evaluated linearization. Since there are fifteen latent coordinates and twelve outputs, the full decomposition also contains three additional latent directions with zero output response.

This separation of measurement-sensitive and null-space directions follows sensitivity-based approaches to parameter identifiability~\cite{Doherty2009Identifiability,Stigter2015Identifiability}.

\paragraph{Rank and conditioning.}

For the adopted model, the first nine singular values, normalized by the largest, are
\begin{equation}
\begin{aligned}
\frac{1}{\sigma_1}
\begin{bmatrix}
\sigma_1 & \cdots & \sigma_9
\end{bmatrix}
=
\bigl[
&1.000,\ 0.751,\ 0.459,\ 0.352,\ 0.236,\\
&0.216,\ 0.172,\ 0.145,\ 0.062
\bigr],
\end{aligned}
\end{equation}
while the remaining three singular values are zero to numerical precision. The mean Jacobian therefore has rank 9 and a six-dimensional null space.

The ratio between the largest and smallest retained singular values is $\sigma_1/\sigma_9=16.1\pm0.1$ across the five seeds. This ratio describes the relative sensitivity of the retained directions. The gap between the ninth and tenth singular values is of the order of $10^{15}$, separating these directions clearly from the numerical null space. The results characterize the evaluated mean Jacobian; estimation accuracy additionally depends on measurement uncertainty and model fidelity.
A $12\times15$ Jacobian necessarily has at least three null-space directions because the number of latent coordinates exceeds the number of outputs. The measured rank of nine therefore gives three additional null-space directions beyond this dimensional minimum. The decoder determines the orientation of the resulting six-dimensional null space and its contributions to the individual latent coordinates.
The stability of the numerical rank was examined by repeating the calculation with finite-difference steps from $10^{-1}$ to $10^{-5}$. Across these calculations, $\sigma_9/\sigma_1$ remains approximately $0.062$, while $\sigma_{10}/\sigma_1$ ranges from $1.4\times10^{-17}$ to $4.4\times10^{-17}$. The rank remains nine at each tested relative tolerance, $10^{-2}$, $10^{-6}$, and $10^{-12}$. These results support the stability of the rank estimate across the tested step sizes and tolerances.

\paragraph{Interpretation of individual coordinates.}

To relate these directions to individual physical quantities, let
\begin{equation}
\mathbf{V}_0
=
\begin{bmatrix}
\mathbf{v}_{10} & \cdots & \mathbf{v}_{15}
\end{bmatrix}
\end{equation}
contain an orthonormal basis of the six-dimensional null space. The null-space share of coordinate $k$ is
\begin{equation}
u_k
=
\left\|\mathbf{V}_0^{\top}\mathbf{e}_k\right\|_2^2
=
\sum_{j=10}^{15}V_{kj}^{2},
\end{equation}
where $\mathbf{e}_k$ is the unit vector associated with that coordinate. This quantity lies between zero and one and is independent of the particular orthonormal basis chosen for the null space. A value near zero indicates that the coordinate has little contribution to the unconstrained directions; a value near one indicates that it lies almost entirely within them. The shares sum to six, the dimension of the null space.

\begin{table}[htbp]
\caption{Sensitivity and local identifiability of the latent coordinates. The sensitivity and loss-gradient norms refer to the initial 16-dimensional representation. The null-space share $u_k$ refers to the adopted 15-dimensional representation and is therefore not defined for the removed coordinate. It quantifies the squared projection of each coordinate onto the null space of the evaluated mean Jacobian.}
\label{tab:observability}
\centering
\begin{tabular}{lcccc}
\toprule
Coordinate & $\|\partial\hat{\mathbf{x}}/\partial z_k\|$
           & $\|\partial\mathcal{L}/\partial z_k\|$
           & $u_k$ & Status \\
\midrule
$\beta_{\mathrm{comb}}$ & $3.77\times10^{-5}$ & $7.86\times10^{-18}$ & --         & removed \\
$\beta_{\mathrm{LPT}}$  & $1.80\times10^{-3}$ & $1.63\times10^{-10}$ & $0.998$    & kept \\
$\beta_{\mathrm{HPT}}$  & $9.52\times10^{-3}$ & $1.04\times10^{-9}$  & $0.993$    & kept \\
$\eta_{\mathrm{LPC}}$   & $1.49\times10^{-2}$ & $7.23\times10^{-10}$ & $0.646$    & kept \\
$\eta_{\mathrm{fan}}$   & $2.03\times10^{-2}$ & $9.73\times10^{-10}$ & $0.354$    & kept \\
$\mathrm{BPR}$         & $2.13\times10^{-2}$ & $1.81\times10^{-9}$  & $0.669$    & kept \\
$W_{\mathrm{mod}}$     & $2.34\times10^{-2}$ & $1.99\times10^{-9}$  & $0.600$    & kept \\
$\eta_{\mathrm{LPT}}$   & $3.57\times10^{-2}$ & $3.01\times10^{-9}$  & $0.101$    & kept \\
$\Pi_{\mathrm{LPT}}$    & $4.30\times10^{-2}$ & $1.87\times10^{-9}$  & $0.427$    & kept \\
$\Pi_{\mathrm{HPT}}$    & $4.71\times10^{-2}$ & $2.42\times10^{-9}$  & $0.427$    & kept \\
$\eta_{\mathrm{HPT}}$   & $4.73\times10^{-2}$ & $4.83\times10^{-9}$  & $0.639$    & kept \\
$\Pi_{\mathrm{HPC}}$    & $7.47\times10^{-2}$ & $3.24\times10^{-9}$  & $<10^{-5}$ & kept \\
$\eta_{\mathrm{HPC}}$   & $8.44\times10^{-2}$ & $3.23\times10^{-9}$  & $<10^{-5}$ & kept \\
$\Pi_{\mathrm{LPC}}$    & $1.03\times10^{-1}$ & $5.40\times10^{-9}$  & $<10^{-5}$ & kept \\
$\Pi_{\mathrm{fan}}$    & $1.22\times10^{-1}$ & $6.31\times10^{-9}$  & $<10^{-5}$ & kept \\
\bottomrule
\end{tabular}
\end{table}

Table~\ref{tab:observability} shows negligible null-space shares for $\Pi_{\mathrm{fan}}$, $\Pi_{\mathrm{LPC}}$, $\Pi_{\mathrm{HPC}}$, and $\eta_{\mathrm{HPC}}$, with $u_k<10^{-5}$. The two retained cooling fractions have $u_k>0.99$, while the remaining coordinates have intermediate values.

These results also explain why sensitivity alone is insufficient to assess identifiability. Although $\eta_{\mathrm{HPT}}$ has a larger Jacobian-column norm than $\eta_{\mathrm{LPT}}$, its null-space share is also larger, approximately $0.64$ compared with $0.10$. Its influence on the measurements is therefore stronger but more closely shared with other latent quantities.

\section{Nominal-Engine Calibration}
\label{app:nominal_calibration}

The physical decoder is constructed from general textbook thermodynamic relations for a two-spool turbofan engine and is not derived from the internal N-CMAPSS simulation model. Its equations describe the main component and gas-path relations required by the proposed architecture, while several details of the high-fidelity simulation used to generate N-CMAPSS are intentionally not represented. A direct agreement between the nominal decoder and the healthy N-CMAPSS engine is therefore not expected.

The fixed quantities collected in $\boldsymbol{\Theta}$ define the nominal engine corresponding to $\mathbf{z}=\mathbf{0}$. Initial values are assigned from conventional turbofan assumptions and engineering ranges. When applied directly to the healthy development data, this nominal configuration gives an overall reconstruction $R^2=-31.03$, reflecting the mismatch between the generic thermodynamic formulation and the specific engine represented in N-CMAPSS. A preliminary calibration is therefore performed to align the nominal reference with the healthy dataset before training the autoencoder \cite{TIAN2022108284}.

\paragraph{Calibration procedure.}
Fifteen fixed quantities are identified using healthy samples from the development units, while all latent modifiers are held at their nominal value. The calibrated parameter vector is obtained from

\begin{equation}
\widehat{\boldsymbol{\Theta}}
=
\arg\min_{\boldsymbol{\Theta}\in[\boldsymbol{\ell},\boldsymbol{u}]}
\sum_{t\in\mathcal{T}_{\mathrm{H}}^{\mathrm{dev}}}
\left\|
\mathcal{P}
\left(
\mathbf{w}_t,
\mathbf{p}_0(\boldsymbol{\Theta});
\boldsymbol{\Theta}
\right)
-
\mathbf{x}_t
\right\|_2^2
+
\lambda_{\Pi}\,\mathcal{L}_{\Pi},
\label{eq:nominal_calibration}
\end{equation}

with

\begin{equation}
\mathcal{L}_{\Pi}
=
\sum_t
\left[
\left(
\frac{\Pi_{\mathrm{HPT}}}{\Pi_{\mathrm{HPT}}^{\max}}-1
\right)_+^2
+
\left(
\frac{\Pi_{\mathrm{LPT}}}{\Pi_{\mathrm{LPT}}^{\max}}-1
\right)_+^2
\right],
\end{equation}

where $\Pi_{\mathrm{HPT}}^{\max}=6$,
$\Pi_{\mathrm{LPT}}^{\max}=8$, and $\lambda_{\Pi}=1$.
The penalty acts only when the corresponding expansion ratio exceeds the prescribed physical range.

The calibration set contains the first $15\%$ of the cycles of each development unit, following the same healthy-window definition used for the operating-condition correction. The resulting pool is subsampled to $20\,000$ observations. Test units are excluded from the calibration. The optimization is performed using trust-region reflective least squares. The effect of this preliminary calibration on the reconstruction of each measured variable is reported in Table~\ref{tab:nominal_calibration_fit}.

\begin{table}[t]
\centering
\caption{Reconstruction $R^2$ of the nominal physical model before and after
calibration on healthy development samples.}
\label{tab:nominal_calibration_fit}
\renewcommand{\arraystretch}{1.05}
\setlength{\tabcolsep}{5pt}
\footnotesize
\begin{tabular}{lrr@{\hspace{12pt}}lrr}
\toprule
\textbf{Variable} & \textbf{Initial} & \textbf{Calibrated} &
\textbf{Variable} & \textbf{Initial} & \textbf{Calibrated} \\
\midrule
$T_{24}$  & $-160.8$ & $0.312$  &
$P_{24}$  & $-121.4$ & $0.775$ \\

$T_{30}$  & $-45.8$  & $-0.218$ &
$P_{s30}$ & $-32.9$  & $0.324$ \\

$T_{48}$  & $-0.661$ & $0.051$  &
$P_{40}$  & $-33.2$  & $0.331$ \\

$T_{50}$  & $-78.5$  & $-1.007$ &
$P_{50}$  & $-11.8$  & $0.465$ \\

$P_{15}$  & $0.486$  & $0.919$  &
$W_f$     & $0.167$  & $0.175$ \\

$P_2$     & $0.9998$ & $0.9998$ &
$P_{21}$  & $0.36945$ & $0.91920$ \\
\midrule
\multicolumn{4}{r}{\textbf{Overall}} &
$\mathbf{-31.03}$ & $\mathbf{0.534}$ \\
\bottomrule
\end{tabular}
\end{table}

The calibration increases the overall nominal-model $R^2$ from $-31.03$ to $0.534$. The remaining mismatch is expected, since a single set of constants within a simplified textbook thermodynamic model is used to represent the healthy behaviour of multiple engines generated by a substantially more detailed simulation environment. The purpose of the calibration is therefore to establish a reasonable reference around which the latent modifiers can evolve. After training, the full physics-constrained autoencoder reaches $R^2=0.9993$, showing that the inferred physical modifiers provide the additional degrees of freedom required to reproduce the observed engine response. The initial values, calibrated values, and optimization bounds of the nominal-engine parameters are reported in Table~\ref{tab:nominal_parameters}.

\begin{table}
\centering
\caption{Initial and calibrated nominal-engine parameters. Parameters marked as
bounded lie at, or within numerical tolerance of, one of the prescribed
optimization bounds.}
\label{tab:nominal_parameters}
\renewcommand{\arraystretch}{1.04}
\setlength{\tabcolsep}{4.5pt}
\footnotesize
\begin{tabular}{lrrlc}
\toprule
\textbf{Parameter} &
\textbf{Initial} &
\textbf{Calibrated} &
\textbf{Bounds} &
\textbf{At bound} \\
\midrule

$\Pi_{\mathrm{fan}}^0$ & $1.500$  & $1.306$  & $[1.10,\,2.20]$  & \\
$\Pi_{\mathrm{LPC}}^0$ & $3.500$  & $1.222$  & $[1.05,\,5.00]$  & \\
$\Pi_{\mathrm{HPC}}^0$ & $11.429$ & $15.594$ & $[5.00,\,25.00]$ & \\
$\mathrm{BPR}^0$       & $9.000$  & $10.235$ & $[3.00,\,15.00]$ & \\

$\eta_{\mathrm{fan}}^0$ & $0.880$ & $0.918$ & $[0.60,\,0.95]$ & \\
$\eta_{\mathrm{LPC}}^0$ & $0.880$ & $0.863$ & $[0.60,\,0.95]$ & \\
$\eta_{\mathrm{HPC}}^0$ & $0.880$ & $0.897$ & $[0.60,\,0.95]$ & \\
$\eta_{\mathrm{HPT}}^0$ & $0.890$ & $0.894$ & $[0.60,\,0.95]$ & \\
$\eta_{\mathrm{LPT}}^0$ & $0.890$ & $\mathbf{0.950}$ &
$[0.60,\,0.95]$ & Yes \\

$\beta_{\mathrm{comb}}^0$ & $0.075$ & $0.084$ &
$[0.02,\,0.20]$ & \\
$\beta_{\mathrm{HPT}}^0$ & $0.047$ & $\mathbf{0.120}$ &
$[0.02,\,0.12]$ & Yes \\
$\beta_{\mathrm{LPT}}^0$ & $0.033$ & $\mathbf{0.060}$ &
$[0.01,\,0.06]$ & Yes \\

$T_{41}^{0}$ [K] & $1100$ & $1360.6$ & $[800,\,1500]$ & \\
$T_{41}'$ [K] & $500$ & $33.2$ & $[0,\,900]$ & \\
$A_{\mathrm{in}}$ [m$^2$] & $8.296$ & $6.259$ &
$[2.00,\,20.00]$ & \\

\bottomrule
\end{tabular}
\end{table}

Three calibrated quantities reach their prescribed upper bounds: $\eta_{\mathrm{LPT}}^0$, $\beta_{\mathrm{HPT}}^0$, and $\beta_{\mathrm{LPT}}^0$. These values are obtained by jointly fitting the nominal decoder to healthy development data and may partly accommodate differences between the simplified thermodynamic formulation and the N-CMAPSS engine. For the two cooling fractions, the weak measurement sensitivity discussed in Appendix~\ref{app:observability} further limits how tightly their values are constrained. The optimizer can therefore drive $\beta_{\mathrm{HPT}}^0$ and the other values toward the boundary of its admissible range while compensating for residual model mismatch elsewhere in the thermodynamic formulation.


\section{Correspondence between N-CMAPSS and Decoder Variables}
\label{app:NCMAPSStable}

The physical decoder follows conventional turbofan station numbering, whereas the measured variables retain the original N-CMAPSS nomenclature. Since the two conventions do not always use the same station labels for equivalent physical quantities, Table~\ref{tab:ncmapss_decoder_mapping} reports their correspondence. This mapping is used throughout the decoder formulation to relate the internal thermodynamic states to the measured and reconstructed channels.

\begin{table*}
\centering
\caption{Correspondence between the N-CMAPSS measurement labels and the internal quantities used by the physical decoder. }
\label{tab:ncmapss_decoder_mapping}
\renewcommand{\arraystretch}{1.08}
\setlength{\tabcolsep}{3pt}
\scriptsize

\begin{tabularx}{\textwidth}{p{3.0cm} p{3.2cm} X}
\toprule
\textbf{N-CMAPSS variable} &
\textbf{Decoder quantity} &
\textbf{Physical interpretation} \\
\midrule

$T_{24}$  & $T_{27}$ & LPC outlet total temperature \\

$T_{30}$  & $T_{3}$  & HPC outlet total temperature \\

$T_{48}$  & $T_{44}$ & HPT outlet total temperature after HPT cooling remix \\

$T_{50}$  & $T_{5}$  & LPT outlet temperature after downstream flow remix \\

\addlinespace

$P_{15}$  & $P_{17}$ & Bypass-duct total pressure \\

$P_{2}$   & $P_{2}$  & Fan-inlet total pressure \\

$P_{21}$  & $P_{13}$ & Fan-outlet total pressure \\

$P_{24}$  & $P_{27}$ & LPC-outlet total pressure \\

$P_{s30}$ & Static $P_{3}$ &
HPC-outlet static pressure, derived from the decoder total pressure $P_{3}$ \\

$P_{40}$  & $P_{41}$ & Combustor-outlet total pressure \\

$P_{50}$  & $P_{7}$ &
LPT-outlet pressure channel, represented in the decoder downstream of the jet-pipe pressure loss \\

\addlinespace

$W_f$ & $W_f$ & Fuel mass flow \\

\midrule

$P_{45}$ & $P_{44}$ &
Inter-turbine total pressure; virtual channel used only in the observability study \\

\bottomrule
\end{tabularx}
\end{table*}

\end{document}